\documentstyle[epsf,CWTWcover,a4]{article}

\newcommand{\Tilde}{{\sc Tilde}}
\newcommand{\Progol}{{\sc Progol}}

\newtheorem{example}{Example}{}

\title{Scaling Up Inductive Logic Programming by Learning From Interpretations\thanks{A version of this report has appeared in Data Mining and Knowledge Discovery 3(1), 1999}}

\author{Hendrik Blockeel \\     
        Luc De Raedt \\          
        Nico Jacobs \\ Bart Demoen}

\date{August 2000}

\CW{297}

\begin{document}

\begin{coverpage}

\begin{abstract}
\begin{quotation}               
 When comparing inductive logic programming (ILP)
 and attribute-value learning techniques, there is a trade-off
 between expressive power and efficiency.
 Inductive logic programming techniques are  typically
 more expressive but also less efficient.
 Therefore, the data sets 
 handled by current inductive logic programming
 systems are small according to general standards
 within the data mining community.
 The main source of inefficiency lies in the assumption that several
 examples may be related to each other, so they cannot be handled
 independently.

 Within the learning from interpretations framework for inductive logic
 programming this
 assumption is unnecessary, which allows to scale up existing ILP
 algorithms.  In this paper we explain this learning setting in the context
 of relational databases.  We relate the setting to propositional data mining
 and to the classical ILP setting, and show that learning from interpretations
 corresponds to learning from multiple relations and thus extends the
 expressiveness of propositional learning, while maintaining its
 efficiency to a large extent (which is not the case in the classical
 ILP setting).

 As a case study, we present two alternative implementations of
 the ILP system Tilde\ (Top-down Induction of Logical DEcision trees): 
 Tilde{\em classic}, which loads all data in main memory, and 
 Tilde{\em LDS}, which loads the examples one by one.
 We experimentally compare the implementations, showing Tilde{\em LDS} can
 handle large data sets (in the order of 100,000 examples or
 100 MB) and indeed scales up linearly in the number of examples.
\end{quotation}
\end{abstract}

\keywords Inductive logic programming, machine learning, data mining.
\AMS \Primary I.2.6,         
     \Secondary I.2.3.       

\end{coverpage}

\section{Introduction}
There is a general trade-off in computer science
between expressive power and efficiency.
Theorem proving in first order logic 
is less efficient but more expressive
than theorem proving in propositional logic.
It is therefore no surprise that 
first order induction techniques (such
as those studied within inductive logic programming)
are less efficient than propositional or attribute-value learning techniques.
On the other hand, inductive logic programming
is able to solve  induction problems
beyond the scope of attribute value learning, cf. (Bratko and Muggleton, 1995).
\nocite{Bratko95-CACM:jrnl}

The computational requirements 
of inductive logic programming systems  
are higher than those of propositional learners due to the following reasons:
first, the space  of clauses considered  by
inductive logic programming systems typically is 
much larger than that of propositional learners and can even be infinite.
Second, testing whether a clause covers an example
is more complex than in attribute value learners.
In attribute value learners an example corresponds
to a single tuple in a relational database,
whereas in inductive logic programming
one example may correspond to multiple tuples
of multiple relations. Therefore, the coverage
test in inductive logic programming 
needs a database system to solve complex queries
or even a theorem prover.
Third,  and this is related to the second point,
in attribute value learning testing whether an example is covered
is done {\em locally}, i.e.
independently of the other examples.
Therefore, even if the data set is huge, a specific
coverage test can be performed efficiently.
This contrasts with the large majority of 
inductive logic programming systems,
such as FOIL (Quinlan, 1990)\nocite{Quinlan90:jrnl} or Progol 
(Muggleton, 1995)\nocite{Muggleton95-NGC:jrnl},
in which coverage is tested {\em globally}, i.e.
to test the coverage of one example the 
whole ensemble of  examples and background theory needs to
be considered\footnote{E.g., testing the coverage of 
$member(a,[b,a])$ may depend on $member(a,[a])$. }.
Global coverage tests are much more expensive
than local ones. Moreover, systems using global coverage tests 
are hard to scale up.  Due to the fact that one single coverage 
test (on one example) typically takes more than constant time
in the size of the database, 
the complexity of induction systems exploiting global coverage tests
will grow more than linearly in the number of examples.

In a more recent setting for inductive logic programming, called learning
from interpretations (De Raedt and D{\v z}eroski, 1994; De Raedt et al., 
1998)\nocite{Deraedt94-AI:jrnl,DeRaedt98:coll}, it is assumed
that each example is a small database (or a part of a
global database), and local coverage tests are performed.
Algorithms using local coverage tests are typically linear in the
number of examples. Furthermore, as each example can be loaded
independently of the other ones, there is no need to use a database
system even when the whole data set cannot be loaded into main memory.

Within the setting of learning from interpretations,
we investigate the issue of scaling up inductive logic programming.
More specifically,
we present two alternative implementations
of the \Tilde\ system (Blockeel and De Raedt, 1998)\nocite{Blockeel98b:jrnl}:
\Tilde{\em classic}, which loads all data in main memory,
and \Tilde{\em LDS}, which loads the examples one by one.
The latter is inspired by the work by Mehta et al. (1996)\nocite{Mehta96:proc}, who propose a level-wise algorithm that needs one pass
through the data per level of the tree it builds.
Furthermore, we experimentally compare the algorithms on large data
sets involving 100,000 examples (in the order of 100 MBytes).  The
experiments clearly show that inductive logic programming systems can
be scaled up to satisfy the standards imposed by the data mining
community.  At the same time, this provides evidence in favor of local
coverage tests (as in learning from interpretations) in inductive
logic programming.

This article is organized as follows.
In Section 2 we introduce the learning from
interpretations setting and relate it to the relational database context.
In Section 3 we introduce first order logical decision trees and discuss
the ILP system \Tilde, which induces such trees.  
Section 4 shows how many propositional techniques can be upgraded to the
learning from interpretations setting (using \Tilde\ as an illustration), and 
discusses why this is much harder for the classical ILP setting. 
Section 5 reports on experiments with \Tilde\ through which we empirically
validate our claims, Section 6 discusses some related work and in
Section 7 we conclude.

\section{The learning setting}

We first introduce the problem specification in a logical 
context, then discuss it in the context of relational databases, and finally
relate it to the standard inductive logic programming setting.

We assume familiarity with Prolog or Datalog (see e.g. (Bratko,
1990)\nocite{Bratko90:other}), 
and relational databases
(see e.g. (Elmasri and Navathe, 1989)\nocite{Elmasri89:other}).

A word on our notation: in logical formulae we will adopt the Prolog
convention that names starting with a capital denote variables, and
names starting with a lowercase character denote constants.

\subsection{Problem specification}

In our framework, each example is a set of facts.
These facts encode the specific properties of the examples
in a database.  Furthermore, 
each example is classified into one of a finite set of
possible classes.  
One may also specify background knowledge in
the form of a Prolog program.

More formally, the problem specification is:
\\
\\
{\bf Given:}
\begin{itemize}
\item a set of classes $C$ (each class label $c$ is a nullary predicate),
\item a set of classified examples $E$ (each element of $E$ is of the 
form $(e,c)$ with $e$ a set of facts and $c$ a class label)
\item and a background theory $B$,
\end{itemize}
{\bf Find:} a hypothesis $H$ (a Prolog program), such that for all $(e,c) \in E $,
\begin{itemize}
\item $H \wedge e \wedge B \models c$, and 
\item $\forall c' \in C-\{c\}: H \wedge e \wedge B \not \models  c'$
\end{itemize}

This setting is known in inductive logic programming under the label
{\em learning from interpretations} (De Raedt and D{\v z}eroski, 1994;
De Raedt, 1997; De Raedt et al., 1998)\nocite{Deraedt94-AI:jrnl,DeRaedt97-AI:jrnl} (an
interpretation is just a set of facts).  Notice that within this
setting, one always learns first order definitions of propositional
predicates (the classes).  An implicit assumption is that the class of
an example depends on that example only, not on any other examples.
This is a reasonable assumption for many classification problems,
though not for all; it precludes, e.g., recursive concept definitions.

\begin{example}

Figure~\ref{bongard} shows a set of pictures each of which is labelled
$\ominus$ or $\oplus$.  The task is to classify new pictures into one of
these classes by looking at the objects in the pictures.
We call this kind of problems Bongard-problems, after
Mikhail Bongard, \nocite{Bongard70:book} who used similar problems for
pattern recognition tests (Bongard, 1970).
 
\begin{figure}
\epsffile{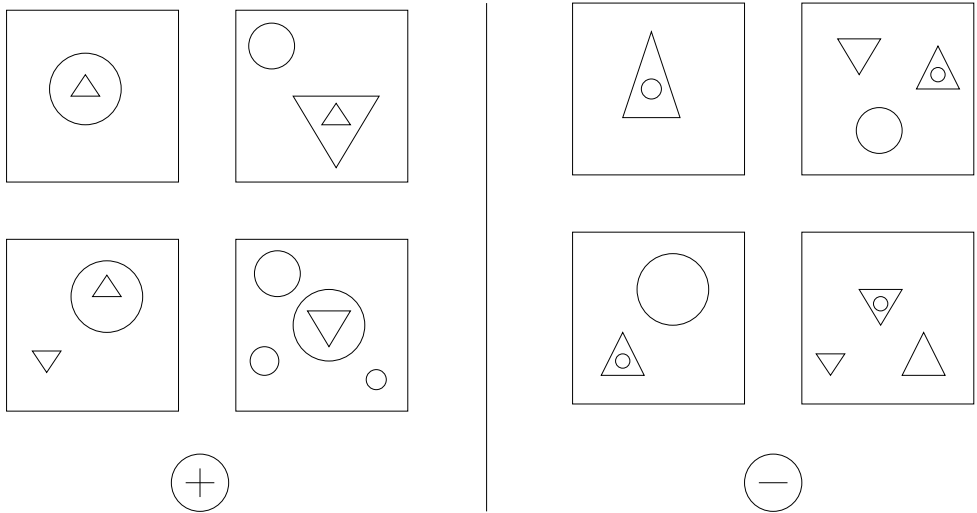}
\caption{Bongard problems}
\label{bongard}
\end{figure}

Assuming we only consider the shape, configuration (pointing upwards
or downwards, for triangles only) and relative position (objects may
be inside other objects) of objects, the pictures in
Figure~\ref{bongard} can be represented as follows:
\begin{flushleft}
Picture 1: \{{\tt circle(o1), triangle(o2), points(o2, up), inside(o2, o1)}\}\\
Picture 2: \{{\tt circle(o3), triangle(o4), points(o4, up), triangle(o5),\\
\qquad \qquad \quad points(o5, down), inside(o4, o5)}\}\\
etc.
\end{flushleft}
(The $o_i$ are constants denoting geometric objects.  The exact names of these
constants are of no importance; they will not be referred to in the first order
hypothesis.)

Background knowledge might be provided to the learner,
e.g., the following definitions could be in the background:
\begin{flushleft}
{\tt doubletriangle(O1,O2) :- triangle(O1), triangle(O2), O1 $\not=$ O2.}\\
{\tt polygon(O) :- triangle(O).}\\
{\tt polygon(O) :- square(O).}
\end{flushleft}

When considering a particular example (e.g. Picture 2)
in conjunction with the background knowledge it is possible
to deduce additional facts in the example.
For instance, in Picture 2, the facts {\tt doubletriangle(o4,o5)}
and {\tt polygon(o4)} hold.

\end{example}

The format of a hypothesis in this setting will be illustrated later.

\subsection{Learning from Multiple Relations}

The learning from interpretations setting, as introduced before, can easily
be related to learning from multiple relations in a relational database.

Typically, each predicate will correspond to one relation in the
relational database.  Each fact in an interpretation is a tuple in the
database, and an interpretation corresponds to a part of the database
(a set of tuples).  Background knowledge can be expressed by means of
views as well as extensional tables.

\begin{example}
For the Bongard example, the following database contains a description
of the first two pictures in Figure~\ref{bongard} (note that an extra
relation CONTAINS is introduced, linking objects to pictures; this relation
was implicit in the previous representation):

\begin{tabbing}
CONTAINS\\
\begin{tabular}{cc}
\hline
{\bf picture} & {\bf object}\\
\hline
1 & o1\\
1 & o2\\
2 & o3\\
2 & o4\\
2 & o5\\
\hline
\end{tabular}
\\
\\
CIRCLE \qquad \= TRIANGLE \qquad \= 
\begin{tabular}{cc}{\bf object} & {\bf direction} \end{tabular} \qquad \= \kill

CIRCLE \> TRIANGLE \> POINTS \> INSIDE\\
\begin{tabular}{c}
\hline
{\bf object}\\
\hline
o1\\
o3\\
\\
\hline
\end{tabular}
\>

\begin{tabular}{c}
\hline
{\bf object}\\
\hline
o2\\
o4\\
o5\\
\hline
\end{tabular}

\>

\begin{tabular}{cc}
\hline
{\bf object} & {\bf direction}\\
\hline
o2 & up\\
o4 & up\\
o5 & down\\
\hline
\end{tabular}

\>

\begin{tabular}{cc}
\hline
{\bf inner} & {\bf outer}\\
\hline
o2 & o1\\
o4 & o5\\
\\
\hline
\end{tabular}
\end{tabbing}

The background knowledge can be defined using views, as follows:
(we are assuming here that a relation SQUARE is also defined)
\begin{verbatim}
DEFINE VIEW doubletriangle AS
SELECT c1.object, c2.object 
FROM contains c1, c2
WHERE c1.object <> c2.object
  AND c1.picture = c2.picture
  AND c1.object IN triangle
  AND c2.object IN triangle;

DEFINE VIEW polygon AS 
SELECT object FROM triangle
UNION
SELECT object FROM square;
\end{verbatim}
\end{example}

In this example the background knowledge is in a sense redundant: it is
computed from the other relations.  This is not necessarily the case.
The following example illustrates this.  It is also a more realistic example
of an application where mining multiple relations is useful.

\begin{example}
Assume that one has a relational database describing molecules.
The molecules themselves are described by listing the atoms and bonds
that occur in them, as well as some properties of the molecule as a whole.
Mendelev's periodic table of elements is a good example of background
knowledge about this domain.

The following tables illustrate what such a chemical database could look like:

\begin{tabbing}
MENDELEV\\
\begin{tabular}{ccccc}
\hline
{\bf number} & {\bf symbol} & {\bf atomic weight} & {\bf electrons in outer layer} &
 \ldots\\
\hline
1 & H & 1.0079 & 1\\
2 & He & 4.0026 & 2 \\
3 & Li & 6.941 & 1\\
4 & Be & 9.0121 & 2\\
5 & B & 10.811 & 3\\
6 & C & 12.011 & 4\\
\ldots & \ldots & \ldots & \ldots & \ldots\\
\hline
\end{tabular}
\\
\\
\begin{tabular}{ccc}{\bf formula} & carbon dioxide & inorganic \end{tabular} \quad \qquad \=
CONTAINS \qquad \kill

MOLECULES \> CONTAINS \\

\begin{tabular}{ccc}
\hline
{\bf formula} & {\bf name} & {\bf class}\\
\hline
$H_2O$ & water & inorganic\\
$CO_2$ & carbon dioxide & inorganic\\
$CO$ & carbon monoxide & inorganic\\
$CH_4$ & methane & organic\\
$CH_3OH$ & methanol & organic\\
\ldots & \ldots & \ldots \\
\hline
\end{tabular}
\>
\begin{tabular}{cc}
\hline
{\bf molecule} & {\bf atom\_id} \\
\hline
$H_2O$ & h2o-1\\
$H_2O$ & h2o-2\\
$H_2O$ & h2o-3\\
$CO_2$ & co2-1\\
$CO_2$ & co2-2\\
\ldots & \ldots \\
\hline
\end{tabular}
\\
\\
{\bf atom\_id} \qquad {\bf element} \qquad \= BONDS \kill
ATOMS \> BONDS \\
\begin{tabular}{cc}
\hline
{\bf atom\_id} & {\bf element}\\
\hline
h2o-1 & H\\
h2o-2 & O\\
h2o-3 & H\\
co2-1 & O\\
\ldots & \ldots \\
\hline
\end{tabular}
\>
\begin{tabular}{ccc}
\hline
{\bf atom\_id1} & {\bf atom\_id2} & {\bf type}\\
\hline
h2o-1 & h2o-2 & single\\
h2o-2 & h2o-3 & single\\
co2-1 & co2-2 & double\\
co2-2 & co2-3 & double\\
\ldots & \ldots & \ldots\\
\hline
\end{tabular}
\end{tabbing}
A possible classification problem here is to classify unseen molecules
into organic and inorganic molecules, based on their chemical structure.

\end{example}

Notice that this representation of examples
and background knowledge upgrades the typical
attribute value learning representation in two respects.
First, in attribute value learning an example corresponds
to a single tuple for a single relation.
Our representation allows for {\em multiple} tuples
in {\em multiple} relations.
Second, it also allows for using background knowledge.

By joining all the relations in a database into one huge relation, one
can of course eliminate the need for learning from multiple relations.
The above example should make clear that in many cases this is not an
option.  The information in Mendelev's table, for instance, would be
duplicated many times.  Moreover, unless a multiple-instance learner is
used (see e.g. (Dietterich et al., 1997))
\nocite{Dietterich96-AI:jrnl} all the atoms a molecule consists of,
together with their properties, have to be stored in one tuple, so
that an indefinite number of attributes is needed; see (De Raedt, 1998) for
a more detailed discussion.\nocite{DeRaedt98:proc}

While mining such a database is not feasible using propositional techniques,
it is feasible using learning from interpretations.  We proceed to show
how a relational database can be converted into a suitable format.

\subsubsection*{Conversion from relational database to interpretations}

Converting a relational database to a set of interpretations can be done
easily and in a semi-automated way, as follows:

\begin{tabbing}
xxx \= xxx \= \kill
Decide which relations are background knowledge.  \\
Let $DB$ be the original database {\em without the background relations}.\\
Choose an attribute in a relation that uniquely identifies the examples.\\
For each value $i$ of that attribute:\\
\> $S$ := set of all tuples in $DB$ containing that value\\
\> repeat\\
\> \> $S$ := $S \, \cup$ set of all tuples in $DB$ referred to by a foreign key in $S$\\
\> until $S$ does not change anymore\\
\> $S_i$ := $S$
\end{tabbing}

The tuples in $S$ are here assumed to be labelled with the name of the
relation they are part of.  A tuple $(attr_1, \ldots, attr_n)$ of a
relation $R$ can trivially be converted to a fact $R(attr_1, \ldots,
attr_n)$.  By doing this conversion for all $S_i$, each $S_i$ becomes
a set of facts describing an individual example $i$.  The extensional
background relations can be converted in the same manner into one set
of facts that forms the background knowledge.  Background relations
defined by views can be converted to equivalent Prolog programs.

The only parts in this conversion process that are hard to automate are
the selection of the background knowledge (typically, one selects those
relations where each tuple can be relevant for many examples) 
and the conversion of view definitions to Prolog programs.  Also, the
user must indicate which attribute should be chosen as an example identifier,
as this depends on the learning task.

\begin{example}
In the chemical database, we choose as example identifier the molecular
formula.  The background knowledge consists of the table MENDELEV.
In order to build a description of $H_2O$, one first collects the tuples
containing $H_2O$; these are present in MOLECULES and CONTAINS.  These
tuples contain references to {\bf atom\_id}'s h2o-$i$, $i=1,2,3$,
so the tuples containing those symbols are also collected (tuples from
ATOMS and BONDS).  These again
refer to the elements $H$ and $O$, which are foreign keys for the MENDELEV
relation.  Since this relation is in the background, no further tuples
are collected.  Converting the tuples to facts, we get the following 
description of $H_2O$:\\
\{molecules('H2O', water, inorganic), contains('H2O', h2o-1), 
contains('H2O', h2o-2), contains('H2O', h2o-3),
atoms(h2o-1, 'H'), atoms(h2o-2, 'O'), atoms(h2o-3, 'H'),
bonds(h2o-1, h2o-2, single), bonds(h2o-2, h2o-3, single)\}
\end{example}

Some variations of this algorithm can be considered.  For instance, when the
example identifier has no meaning except that it 
identifies the example (as the picture numbers 1 and 2 for the Bongard
example), this attribute can be left out from the example description.

The key notion in this conversion process is {\em localization of information}.
It is assumed that for each example only a relatively small part of the
database is relevant, and that this part can be localized and extracted.
From now on, we will refer to this assumption as the {\em locality assumption}.

\subsection{The standard ILP setting}

We now briefly discuss the standard ILP setting and how it differs
from our setting.  For a more thorough discussion of different ILP
settings and the relationships among them we refer to (De Raedt,
1997).\nocite{DeRaedt97-AI:jrnl}

The standard ILP setting (also known as {\em learning from entailment\/}) 
is usually formulated as follows:

\begin{flushleft}
{\bf Given:}
\begin{itemize}
\item a set of positive examples $E^+$ and a set of negative examples $E^-$
\item and a background theory $B$,
\end{itemize}
{\bf Find:} a hypothesis $H$ (a Prolog program), such that 
\begin{itemize}
\item $\forall e \in E^+ : H \wedge B \models e$, and 
\item $\forall e \in E^- : H \wedge B \not \models e$
\end{itemize}
\end{flushleft}

Note that in this setting, an example $e$ is a fact (or clause) that is to be
explained by $H \land B$, while in the learning from interpretations
setting a {\em property} of the example (its class) is to be explained
by $H \land B \land e$.  Thus, the latter setting explicitates the 
separation between example-specific information and general background 
information.

The problem specification as given above is natural for the standard ILP
setting, where one could, for instance, give the
following examples for the predicate {\tt member}:
\begin{verbatim}
+ : member(a, [a,b,c]).
+ : member(d, [e,d,c,b]).
+ : member(d, [d,c,b]).
- : member(b, [a,c,d]).
- : member(a, []).
- : member(d, [c,b]).
\end{verbatim}
and expect the ILP system to come up with the following definition:
\begin{verbatim}
member(X, [X|Y]).
member(X, [Y|Z]) :- member(X,Z).
\end{verbatim}

Note that the class of an example (i.e., its truth value) now depends on 
the class of other examples;
e.g., the class of {\tt member(d, [e,d,c,b])} depends on the class 
of {\tt member(d, [d,c,b])}, which is a different example.
Because of this property, it is in general not possible to find a small subset
of the database that is relevant for a single example, i.e., local coverage
tests cannot be used.  Results from computational learning theory confirm
that learning hypotheses in this setting generally is intractable (see
e.g.\ (D{\v z}eroski et al., 1992; Cohen, 1995; Cohen and Page, 1995)\nocite{Dzeroski92-COLT:proc,Cohen95-JAIR-2:jrnl,Cohen95-NGC:jrnl}).

Since in learning from interpretations the class of an example is assumed
to be independent of other examples, this setting is less powerful than
the standard ILP setting (e.g., for what concerns recursion).  With this 
loss of power comes a gain in efficiency, through local coverage tests.
The interesting point is that the full power of standard ILP
is not used for most practical applications, and learning from 
interpretations usually turns out to be sufficient for practical applications,
see e.g. the proceedings of the ILP workshops and conferences of the
last few years (De Raedt, 1996; Muggleton, 1997; 
Lavra{\v c} and D{\v z}eroski, 1997; Page, 1998).\nocite{DeRaedt96:book,ILP96:other,ILP97:other,ILP98:other}

\section{\Tilde: Induction of First-Order Logical Decision Trees}

In this section, we discuss one specific ILP system that learns from 
interpretations, called \Tilde\ (which stands for Top-down Induction of Logical
DEcision trees).  This system will be used to illustrate
the topics discussed in the following sections.

We first introduce the hypothesis representation formalism used by \Tilde,
then discuss an algorithm for the induction of hypotheses in this formalism.

\subsection{First order logical decision trees}

We will use first order logical decision  trees
for representing hypotheses. 
 These are an upgrade of the well-known propositional decision trees to 
first order learning.

A first order logical decision tree (FOLDT) is a binary decision
tree in which
\begin{itemize}
\item the nodes of the tree contain a conjunction of literals
\item different nodes may share variables, under the following restriction:
a variable that is introduced in a node (which means that it does not occur
in higher nodes) must not occur in the right branch of that node.  The
need for this restriction follows from the semantics of the
tree.  A variable $X$ that is introduced in a node, is quantified existentially
within the conjunction of that node.
The right subtree is only relevant when the
conjunction fails (``there is no such $X$''), in which case
further reference to $X$ is meaningless.
\end{itemize}                                                        
 
\begin{figure}
\epsffile{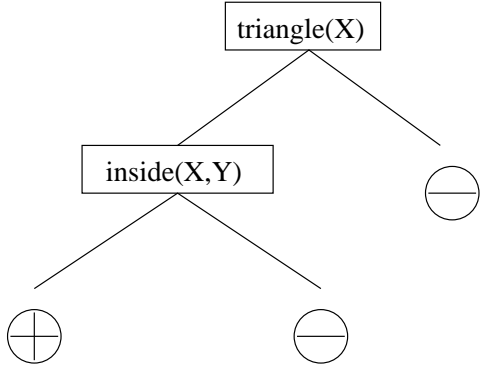}
\caption{A first order logical decision tree that allows to discriminate
the two classes for the Bongard problem shown in Figure~\ref{bongard}.}
\label{bongtree}
\end{figure}

An example of such a tree is shown in Figure~\ref{bongtree}.  

First order logical decision trees can be converted to normal logic programs
(i.e. logic programs that allow negated literals in the body of a clause)
and to Prolog programs.  In the latter case the Prolog program represents
a first order decision list, i.e. an ordered set of rules where a rule
is only relevant if none of the rules before it succeed.  Each clause in such
a Prolog program ends with a cut.  We refer to (Blockeel and De Raedt, 1998) 
\nocite{Blockeel98b:jrnl} for
more information on the relationship between first order decision trees,
first order decision lists and logic programs.

The Prolog program equivalent to the tree in Figure~\ref{bongtree} 
is\footnote{The Prolog program entails {\tt class($c$)} instead of $c$,
in order to ensure that the cuts have the intended meaning; this is a merely
syntactical difference with the original task formulation.}
\begin{verbatim}
class(pos) :- triangle(X), inside(X,Y), !.
class(neg) :- triangle(X), !.
class(neg).
\end{verbatim}

Figure~\ref{classify} shows how to use FOLDTs for classification.
We use the following notation: a tree $T$ is either a leaf with class $c$,
in which case we write $T = \mbox{\bf leaf}(c)$, or it is an
internal node with conjunction {\em conj}, left branch {\em left} and right 
branch {\em right}, in which case we write $T$ = {\bf inode}({\em conj, left, 
right\/}).
 
Because an example $e$ is a Prolog program, a test in a node corresponds to
checking whether a query $\leftarrow C$ succeeds in $e \wedge B$
(with $B$ the background knowledge).  
Note that it is not sufficient to use for $C$ the conjunction {\em conj}
in the node itself.  Since {\em conj} may share variables with nodes higher
in the tree, $C$ consists of several conjunctions that occur in the
path from the root to the current node.  More specifically, $C$ is of
the form $Q \land conj$, where $Q$ is the conjunction of all the 
conditions that occur in those nodes on the path from the root to this node
where the left branch was chosen.  We call $\leftarrow Q$ the {\em associated
query} of the node.

When an example is
sorted to the left, $Q$ is updated by adding {\em conj} to it.
When sorting an example to the right, $Q$ need not be updated: a failed
test never introduces new variables.  E.g., if in Figure~\ref{bongtree}
an example is sorted down the tree, in the node containing
{\tt inside(X,Y)} the correct test is {\tt triangle(X), inside(X,Y)};
it is not correct to test {\tt inside(X,Y)} on its own.

\begin{figure}
\begin{tabbing}
xxx \= xxx \= xxx \= \kill
{\bf procedure} classify($e$ : {\bf example}) {\bf returns class}:\\
\> $Q := true$\\
\> $N := $ root\\
\> {\bf while} $N \not= \mbox{\bf leaf}(c)$ {\bf do}\\
\> \> {\bf let} $N = \mbox{\bf inode}(conj,left,right)$\\
\> \> {\bf if} $Q \land conj$ succeeds in $e \land B$\\
\> \> {\bf then} $Q := Q \land conj$\\
\> \> \> \, $N := left$\\
\> \> {\bf else} $N := right$\\
\> {\bf return} $c$
\end{tabbing}
\caption{Classification of an example using an FOLDT (with background
knowledge $B$)}
\label{classify}
\end{figure}

\subsection{The \Tilde\ system}

\begin{figure}
\begin{tabbing}
xxx \= xxx \= \kill
{\bf procedure} buildtree($T$: {\bf tree}, $E$: {\bf set of examples}, $Q$: {\bf query}):\\
\> $\leftarrow Q_b$ := element of $\rho(\leftarrow Q)$ with highest gain (or gain ratio)\\
\> {\bf if} $\leftarrow Q_b$ is not good  {\em /* e.g. does not yield any gain at all */}\\
\> {\bf then} $T$ := {\bf leaf}(majority\_class(E))\\
\> {\bf else}\\
\> \> {\em conj} := $Q_b - Q$\\
\> \> $E_1$ := $\{ e \in E | \leftarrow Q_b \mbox{ succeeds in } e \land B \}$\\
\> \> $E_2$ := $\{ e \in E | \leftarrow Q_b \mbox{ fails in } e \land B \}$\\
\> \> buildtree({\em left}, $E_1$, $Q_b$)\\
\> \> buildtree({\em right}, $E_2$, $Q$)\\
\> \> $T$ := {\bf inode}({\em conj}, {\em left}, {\em right})\\
\\
{\bf procedure} \Tilde($T$: {\bf tree}, $E$: {\bf set of examples}):\\
\> buildtree($T$, $E$, true)
\end{tabbing}
\caption{Algorithm for first-order logical decision tree induction}
\label{atilde}
\end{figure}

First order logical decision trees can be induced in very much the same
manner as propositional decision trees.  The generic algorithm for this
is usually referred to as TDIDT: top-down induction of decision trees.
Examples of systems using this approach are C4.5 (Quinlan, 1993a) 
\nocite{Quinlan93:other} and CART (Breiman et al., 1984)\nocite{Breiman84:other}.

The algorithm we use for inducing first order decision trees is shown
in Figure~\ref{atilde}.  The \Tilde\ system (Blockeel and De Raedt, 1998)
\nocite{Blockeel98b:jrnl} is
an implementation of this algorithm that is based on
C4.5.  It uses the same heuristics, the same post-pruning algorithm, etc.
 
The main point where our algorithm differs from C4.5 is in the
computation of the set of tests to be considered at a node.  C4.5 only
considers tests comparing an attribute with a value.
\Tilde, on the other hand, generates possible tests by means of a 
user-defined refinement operator.  Roughly, this operator specifies, given
the associated query of a node, which literals or conjunctions can be added
to the query.

More specifically, the refinement operator is a refinement operator 
under $\theta$-sub\-sumption (Plotkin, 1970; Muggleton and De Raedt, 1994).
\nocite{Plotkin70:coll,Muggleton94-JLP:jrnl}
Such an operator $\rho$ maps clauses onto sets of clauses, such that for
any clause $c$ and $\forall c' \in \rho(c)$, $c$ $\theta$-subsumes
$c'$.  A clause $c_1$ $\theta$-subsumes another clause $c_2$ if and only if
there exists a variable substitution $\theta$ such that $c_1\theta
\subseteq c_2$.
The operator could for instance add literals to the clause, or unify
several variables in it.  The use of such refinement operators is
standard practice in ILP.

In order to refine a node with associated query $\leftarrow Q$,
\Tilde\ computes $\rho(\leftarrow Q)$ and chooses the 
query $\leftarrow Q_b \in \rho(\leftarrow Q)$ that
results in the best split.  The best
  split is the one that maximizes a certain quality criterion; in the
case of \Tilde\ this is by default the information gain ratio, as defined by 
Quinlan (1993a).\nocite{Quinlan93:other}
The conjunction put in the node consists
of $Q_b - Q$, i.e., the literals that have been added to $Q$ in order
to produce $Q_b$.

\begin{example}
  Consider the tree in Figure~\ref{bongtree}.  Assuming that
  the root node has already been filled in with the test {\tt
    triangle(X)}, how does \Tilde\ process the left child of it?
  This child has as associated query $\leftarrow${\tt triangle(X)}.
  \Tilde\ now generates $\rho(\leftarrow {\tt triangle(X)})$.
  According to the language bias specified by the user (see below), a
  possible result could be (we use semicolons to separate the elements
  of $\rho$, as the comma denotes a conjunction in Prolog)
\begin{tabbing}
$\rho(\leftarrow {\tt triangle(X)})$ = \{ \= $\leftarrow$ {\tt triangle(X), inside(X,Y)};\\
\> $\leftarrow$ {\tt triangle(X), inside(Y,X)};\\
\> $\leftarrow$ {\tt triangle(X), square(Y)};\\
\> $\leftarrow$ {\tt triangle(X), circle(Y)} \}
\end{tabbing}
Assuming the best of these refinements is 
$Q_b$ = {\tt triangle(X), inside(X,Y)} 
the conjunction put in the node is $Q_b - Q$ = {\tt inside(X,Y)}.
\end{example}

\subsubsection*{Language bias}

While propositional systems usually have a fixed language bias,
most ILP systems make use of a language bias that has been
provided by the user.  The language bias specifies what kind of hypotheses
are allowed; in the case of \Tilde: what kind of literals or conjunctions
of literals can be put in the nodes of the tree.  This bias follows from 
the refinement operator, so it is sufficient to specify the latter.
The specific refinement operator that is to be used is defined by the
user in a \Progol-like manner (Muggleton, 1995). 
\nocite{Muggleton95-NGC:jrnl}  A set of
facts of the form {\tt rmode}($n$:~{\em conjunction\/}) is
provided, indicating which conjunctions can be added to a query, the
maximal number of times the conjunction can be added (i.e. the maximal
number of times it can occur in any path from root to leaf, $n$), and
the modes and types of its variables.

To illustrate this, we return to the example of the Bongard problems.
A suitable refinement operator definition in this case would be
\begin{verbatim}
rmode(5: triangle(+-V)).
rmode(5: square(+-V)).
rmode(5: circle(+-V)).
rmode(5: inside(+V,+-W)).
rmode(5: inside(-V,+W)).
rmode(5: config(+V,up)).
rmode(5: config(+V,down)).
\end{verbatim}

The mode of an argument is indicated by a $+$, $-$ or $+-$ sign before
a variable.  $+$ stands for input: the variable should already occur
in the associated query of the node where the test is put.  $-$ stands for
output: the variable has to be one that does not occur yet.  $+-$ means that 
the argument can be both input and output; i.e. the variable can be a
new one or an already existing one. 
Note that the names of the variables in the {\tt rmode} facts are formal
names; when the literal is added to a clause actual variable names are
substituted
for them.  Also note that a literal can have multiple modes, e.g. the
above facts specify that at least one of the two arguments of {\tt inside} 
has to be input.

This {\tt rmode} definition tells \Tilde\ that a test in a node may 
consist of checking whether an object that has already been referred to 
has a certain shape (e.g. {\tt triangle(X)} with {\tt X} an already existing
variable), checking whether there exists an object with a certain
shape in the picture (e.g. {\tt triangle(Y)} with {\tt Y} not occurring in the
associated query), testing the configuration ({\tt up} or {\tt down}) 
of a certain object, and so on.  At most 5 literals of a certain
type can occur on any path from root to leaf (this is indicated by the 5
in the {\tt rmode} facts).

The decision tree shown in Figure~\ref{bongtree} conforms to this
specification.  When \Tilde\ builds this tree, in the root node only
the tests {\tt triangle(X)}, {\tt square(X)} and {\tt circle(X)} are
considered, because each other test requires some variable to occur in the
associated query of the node (which for the root node is {\em true}).
The left child node of the root has as associated query
$\leftarrow$ {\tt triangle(X)}, which contains one variable {\tt X}, hence
the tests that are considered for this node are:
\begin{verbatim}
triangle(X)      triangle(Y)       inside(X,Y)       points(X,up)
square(X)        square(Y)         inside(Y,X)       points(X,down)
circle(X)        circle(Y)          
\end{verbatim}
Assuming that {\tt inside(X,Y)} yields the best 
split, this literal is put in the node.  

In addition to {\tt rmode}s, so-called lookahead specifications 
can be provided.  These allow \Tilde\ to perform several successive
refinement steps at once.  This alleviates the well-known problem in ILP 
(see e.g. (Quinlan, 1993b)\nocite{Quinlan93:proc}) that
a refinement may not yield any gain, but may introduce new variables that
are crucial for classification.  By performing successive refinement steps
at once, \Tilde\ can look ahead in the refinement lattice and discover such
situations.

For instance, {\tt lookahead(triangle(T), points(T,up))} specifies that
whenever the literal {\tt triangle(T)} is considered as possible addition
to the current associated query, 
additional refinement by adding {\tt points(T,up)} should be tried in
the same refinement step.  Thus, both {\tt triangle(T)} and
{\tt triangle(T), points(T,up)} would be considered as possible addition.
This is useful because normally \Tilde\ can construct the test
{\tt triangle(T), points(T,up)} only by first putting {\tt triangle(T)}
in the node, then putting {\tt points(T,up)} in its left child node.
But if {\tt triangle(X)} already occurs in the associated
query, then {\tt triangle(T)} cannot yield any gain (if you already know that
there is a triangle, the question ``is there a triangle'' will not give you
new information) and hence would never be selected, and this would prevent
{\tt points(T,up)} from being added as well.

This lookahead method is very similar to lookahead methods that have
been proposed for propositional decision tree learners.  While for 
propositional systems the advantage of lookahead is generally considered
to be marginal, it is much greater in ILP because of the occurrence of
variables.

We finally mention that \Tilde\ handles numerical data by means of a
discretization algorithm that is based on Fayyad and Irani's (1993)
\nocite{Fayyad93:proc} and Dougherty et al.'s (1995)
\nocite{Dougherty95:proc} work, but extends it to first order logic
(Van Laer et al., 1997)\nocite{VanLaer97:proc}.  The algorithm 
accepts input of the form
{\tt discretize(Query, Var)}, with {\tt Var} a variable occurring in
{\tt Query}.  It runs {\tt Query} in all the examples, collecting all
instantiations of {\tt Var} that can be found, and finally generates
discretization thresholds based on this set of instantiations.  Since
this discretization procedure is not crucial to this paper, we refer
to (Van Laer et al., 1997; Blockeel and De Raedt, 1997) 
\nocite{VanLaer97:proc,Blockeel97:proc} for more details.

\subsubsection*{Input Format}

A data set is presented to \Tilde\ in the form of a set of interpretations.
Each interpretation consists of a number of Prolog facts, surrounded by
a {\tt begin} and {\tt end} line.  The background knowledge is simply
a Prolog program.  Examples of this will be shown in 
Section~\ref{experiments:sec}.

\subsubsection*{Applications of \Tilde}

Although the above discussion of \Tilde\ takes the viewpoint of induction
of classifiers, the use of first order logical decision trees is not
limited to classification.  Numerical predictions can be made by storing
numbers instead of classes in the leaves; such trees are usually called
regression trees.  Another task that is important
for data mining, is clustering.  Induction of cluster hierarchies can also
be done using a TDIDT approach, as is explained in
(Blockeel et al., 1998)\nocite{Blockeel98:proc}.

It should be clear, therefore, that the techniques that will be
described later in this text should not be seen as specific for the
classification context.  They have a much broader application domain.

\section{Upgrading Propositional KDD Techniques for \Tilde}
 
In this section we discuss how existing propositional KDD techniques
can be upgraded to first order learning in our setting.  The \Tilde\ 
system will serve as a case study here.  Indeed, all of the techniques
proposed below (except sampling) have been implemented in \Tilde.  We
stress, however, that the methodology of upgrading KDD techniques is
not specific for \Tilde, nor for induction of decision trees.  It can
also be used for rule induction, discovery of association rules, and
other kinds of discovery.  Systems such as {\sc Claudien} (De Raedt and
Dehaspe, 1997),\nocite{DeRaedt97-ML:jrnl} ICL (De Raedt and Van Laer, 1995)
\nocite{Deraedt95-ALT:proc} and {\sc Warmr} (Dehaspe and De Raedt, 1997)
\nocite{Dehaspe97a:proc} are illustrations of this.  Both learn from
interpretations and upgrade propositional techniques.  ICL learns
first order rule sets, upgrading the techniques used in CN2, and {\sc
  Warmr} learns a first order equivalent of association rules
(``association rules over multiple relations'').  {\sc Warmr} has been
designed specifically for large databases and employs an efficient
algorithm that is an upgrade of {\sc Apriori} (Agrawal et al., 1996).
\nocite{Agrawal96:coll}

\subsection{Different Implementations of \Tilde}

We discuss two different implementations of \Tilde: one is a straightforward
implementation, following closely the TDIDT algorithm.  The other is a
more sophisticated implementation that aims specifically at handling large
data sets; it is based on work by Mehta et al. (1996) \nocite{Mehta96:proc}, and as such is our first example of how propositional techniques can be
upgraded.

\subsubsection{A straightforward implementation: \Tilde{\it classic}}

The original \Tilde\ implementation, which we will refer to as
\Tilde{\em classic}, is based on the algorithm shown in
Figure~\ref{atilde}.  This is the most straightforward way of implementing
TDIDT.

Noteworthy characteristics are that the tree is
built depth-first, and that the best test is chosen by enumerating
the possible tests and for each test computing its quality (to this aim the
test needs to be evaluated on every single example), as is shown in
Figure~\ref{besttest}.  This algorithm should be seen as a detailed description
of line 6 in Figure~\ref{atilde}.
\begin{figure}
\begin{tabbing}
xxx \= xxx \= \kill
{\bf for each} refinement $\leftarrow Q_i$:\\
\> {\em /* counter[true] and counter[false] are class distributions,}\\
\> \> {\em i.e. arrays mapping classes onto their frequencies */}\\
\> {\bf for each} class $c$ : counter[true][$c$] := 0, counter[false][$c$] := 0\\
\> {\bf for each} example $e$: \\
\> \> {\bf if} $\leftarrow Q_i$ succeeds in $e$\\
\> \> {\bf then} increase counter[true][class(e)] by 1\\
\> \> {\bf else} increase counter[false][class(e)] by 1\\
\> $s_i$ := weighted\_average\_class\_entropy(counter[true], counter[false])\\
$Q_b$ := that $Q_i$ for which $s_i$ is minimal {\em /* highest gain */}
\end{tabbing}
\caption{Computation of the best test $Q_b$ in \Tilde{\it classic}.}
\label{besttest}
\end{figure}

Note that with this implementation, it is crucial that fetching an example
from the database in order to query it is done as efficiently as possible,
because this operation is inside the innermost loop.
For this reason, \Tilde{\em classic} loads all data into main memory when
it starts up.  Localization is then achieved by using the module system of
the Prolog engine in which \Tilde\ runs.  Each example is loaded into a
different module, and accessing an example is done by changing the
currently active module, which is a very cheap operation.
One could also load all the examples into one module; no example selection
is necessary then, and all data can always be accessed directly.  The 
disadvantage is that the relevant data needs to be looked up in a large set
of data, so that a good indexing scheme is necessary in order to make this
approach efficient.  We will return to this in the section on experiments.

We point out that, when examples are loaded into different modules,
\Tilde{\em classic} partially exploits the locality assumption (in
that it handles each individual example independently from the others,
but still loads all the examples in main memory).  It does not
exploit this assumption at all when all the examples are loaded into one
module.

\subsubsection{A more sophisticated implementation: \Tilde{\it LDS}}

Mehta et al. (1996) \nocite{Mehta96:proc} proposed an alternative
implementation of TDIDT that is oriented towards mining large
databases.  With their approach, the database is accessed
less intensively, which results in an important efficiency gain.
We have adopted this approach for an alternative implementation of
\Tilde, which we call \Tilde{\em LDS} ({\em LDS} stands for {\em Large
  Data Sets\/}).

The alternative algorithm is shown in Figure~\ref{alds}.  It differs
from \Tilde{\em classic} in that the tree is now built breadth-first,
and examples are loaded into main memory one at a time.

\begin{figure}
\begin{tabbing}
xxx \= xxx \= xxx \= xxx \= \kill
{\bf procedure} \Tilde{\em LDS}:\\
\> S := \{root\}\\
\> {\bf while} $S \not= \phi$ {\bf do}\\
\> \> {\em /* add one level to the tree */}\\
\> \> {\bf for each} example $e$ that is not covered by a leaf node:\\
\> \> \> load $e$\\
\> \> \> $N$ := the node in $S$ that covers $e$\\
\> \> \> $\leftarrow Q$ := associated\_query($N$)\\
\> \> \> {\bf for each} refinement $\leftarrow Q_i$ of $\leftarrow Q$:\\
\> \> \> \> {\bf if} $\leftarrow Q_i$ succeeds in $e$\\
\> \> \> \> {\bf then} increase counter[$N$,$i$,true][class($e$)] by 1\\
\> \> \> \> {\bf else} increase counter[$N$,$i$,false][class($e$)] by 1\\
\> \> {\bf for each} node $N \in S$ : \\
\> \> \> remove $N$ from $S$\\
\> \> \> $\leftarrow Q_b$ := best\_test($N$)\\
\> \> \> {\bf if} $\leftarrow Q_b$ is not good\\
\> \> \> {\bf then} $N$ := {\bf leaf}(majority\_class($N$))\\
\> \> \> {\bf else} \\
\> \> \> \> $\leftarrow Q$ := associated\_query($N$)\\
\> \> \> \> {\em conj} := $Q_b - Q$\\
\> \> \> \> $N$ := {\bf inode}({\em conj}, {\em left}, {\em right}\/)\\
\> \> \> \> add {\em left} and {\em right} to $S$\\
\\
{\bf function} best\_test($N$: {\bf node}) {\bf returns} query: \\
\> $\leftarrow Q$ := associated\_query($N$)\\
\> {\bf for each} refinement $\leftarrow Q_i$ of $\leftarrow Q$:\\
\> \> $CD_l$ := counter[$N$,$i$,true]\\
\> \> $CD_r$ := counter[$N$,$i$,false]\\
\> \> $s_i$ := weighted\_average\_class\_entropy($CD_l$, $CD_r$)\\
\> $Q_b$ := that $Q_i$ for which $s_i$ is minimal\\
\> {\bf return} $\leftarrow Q_b$
\end{tabbing}
\caption{The \Tilde{\it LDS} algorithm}
\label{alds}
\end{figure}

The algorithm works level-wise. Each iteration through the {\bf while}
loop will expand one level of the decision tree. $S$ contains all
nodes at the current level of the decision tree.  To expand this
level, the algorithm considers all nodes $N$ in $S$.  For each node
and for each refinement in that node, a separate counter (to compute
class distributions) is kept.  The algorithms makes one pass through
the data, during which for each example that belongs to a non-leaf
node $N$ it tests all refinements for $N$ on the example and updates
the corresponding counters.

Note that while for \Tilde{\em classic} the example loop was inside the
refinement loop, the opposite is true now.  This minimizes the number
of times a new example must be loaded, which is an expensive operation
(in contrast with the previous approach where all examples were in
main memory and examples only had to be ``selected'' in order to
access them, examples are now loaded from disk).  In the current
implementation each example needs to be loaded at most once per
level of the tree (``at most'' because once it is in a leaf it need
not be loaded anymore), hence the total number of passes through the data
file is equal to the depth of the tree, which is the same as
was obtained for propositional learning algorithms (Mehta et al., 1996).
\nocite{Mehta96:proc}

The disadvantage of this algorithm is that a four-dimensional array of
counters needs to be stored instead of a two-dimensional one (as in
\Tilde{\em classic}), because different counters are kept for each
node and for each refinement.  

Care has been taken to implement \Tilde{\em LDS} in such a way that
the size of the data set that can be handled is not restricted
by internal memory (in contrast to \Tilde{\em classic}).  Whenever
information needs to be stored the size of which depends on the size
of the data set, this information is stored on disk.\footnote{The
  results of all queries for each example are stored in this manner,
  so that when the best query is chosen after one pass through the
  data, these results can be retrieved from the auxiliary file, avoiding
  a second pass through the data.}  When
processing a certain level of the tree, the space complexity
of \Tilde{\em LDS} therefore contains a component $O(r \cdot n)$ with $n$ 
the number of nodes on that level and $r$ the (average) number of 
refinements of those nodes (because counters are kept for each refinement
in each node), but is constant in the number of examples.  
This contrasts with \Tilde{\em classic} where space complexity contains
a component $O(m)$ with $m$ the number of examples (because all examples
are loaded at once).

While memory now restricts the number of refinements that can be
considered in each node and the maximal size of the tree, this
restriction is unimportant in practice, as the number of refinements
and the tree size are usually much smaller than the upper bounds
imposed by the available memory.  Therefore \Tilde{\em LDS} typically
consumes less memory than \Tilde{\em classic}, and may be preferable
even when the whole data set can be loaded into main memory.

\subsection{Sampling}

While the above implementation is one step towards handling large data
sets, there will always be data sets that are too large to handle.  An
approach that is often taken by data mining systems when there are too
many examples, is to select a sample from the data and learn from that
sample.  Such techniques are incorporated in e.g. C4.5
(Quinlan, 1993a)\nocite{Quinlan93:other} and CART (Breiman et al., 
1984)\nocite{Breiman84:other}.

In the standard ILP context there are some difficulties with sampling, which
can be ascribed to the lack of a locality assumption.  When one example
contains information that is relevant for another example, either both examples
have to be included together in the sample, or none of them should.  Otherwise,
one obtains a sample in which some examples have an incomplete description
(and hence are noisy).  It is even possible that no good sample can be drawn
because all the examples are related to one another.
To the best of our knowledge sampling has received little attention 
inside ILP, as is also noted by F{\"u}rnkranz (1997a) and Srinivasan 
(1998). \nocite{Fuernkranz97b:proc}

If the locality assumption can be made, such sampling problems do not occur.  
Picking individual examples from the population in a random fashion, 
independently from one another, is sufficient to create a good sample.

Automatic sampling has not been included in the current \Tilde\ 
implementations.  We do not give this high priority because \Tilde\ learns
from a flat file of data which is produced by extracting
information from a database and putting related information together (as
explained earlier in this text).  Sampling should be done at the level of
the extraction of information, not by \Tilde\ itself.  It is rather
inefficient to convert the whole database into a flat file and then use
only a part of that file, instead of only converting the part of the
database that will be used.  

We do not present experiments with sampling, as the effect
of sampling in data mining is out of the scope of this paper; instead we refer
to the already existing studies on this subject (see e.g.
(Muggleton, 1993; F{\"u}rnkranz, 1997b; Srinivasan, 1999)).\nocite{Muggleton93-ALT:proc,Fuernkranz97:proc,Srinivasan99:jrnl}

\subsection{Internal Validation}

Internal validation means that a part of the learning set (the
validation set) is kept apart for validation purposes, and the rest is
used as the training set for building the hypothesis.  Such a
methodology is often followed for tuning parameters of a system or for
pruning.  Similar to sampling, partitioning the learning set is easy
if the locality assumption holds, otherwise it may be hard; hence learning
from interpretations makes it easier to incorporate validation
based techniques in an ILP system.

\subsection{Scalability}

\label{scalability:sec}

De Raedt and D{\v z}eroski (1994) \nocite{Deraedt94-AI:jrnl} have
shown that in the learning from interpretations setting, learning
first-order clausal theories is tractable.  More specifically, given
fixed bounds on the maximal length of clauses and the maximal arity of
literals, such theories are polynomial-sample polynomial-time
PAC-learnable.  This positive result is related directly to the
learning from interpretations setting.

Quinlan (1986) \nocite{Quinlan86:jrnl} has shown that induction of decision
trees has time complexity $O(a \cdot N \cdot n)$ where $a$ is the number of
attributes of each example, $N$ is the number of examples and $n$ is
the number of nodes in the tree.  Since \Tilde\ uses basically the
same algorithm as Quinlan, it inherits the linearity in the number of
examples and in the number of nodes.  The main difference between
\Tilde\ and C4.5, as we already noted, is the generation of tests in a
node.  

The number of tests to be considered in a node depends on the
refinement operator.  There is no theoretical bound on this, as it is
possible to define refinement operators that cause an infinite
branching factor.  In practice, useful refinement operators always
generate a finite number of refinements, but even then this number may
not be bounded: the number of refinements typically increases with the
length of the associated query of the node.  Also, the time for
performing one single test on a single example depends on the
complexity of that test (it is in the worst case exponential in the
length of the conjunction).

Thus, we can say that induction of first order decision trees has time
complexity $O(N \cdot n \cdot t \cdot c)$ with $t$ the average number
of tests performed in each node and $c$ the average time complexity of
performing one test for one example, if those averages exist.
If one is willing to accept an upper bound on the complexity of the
theory that is to be learned (which was done for the PAC-learning
results) and defines a finite refinement operator, both the complexity
of performing a single test on a single example and the number of
tests are bounded and the averages do exist.

Our main conclusion from this is that the time complexity of \Tilde\ 
is linear in the number of examples.  This is a stronger claim than
can be made for the standard ILP setting.  The time complexity
also depends on the global complexity of the theory and the branching
factor of the refinement operator, which are domain-dependent parameters.

\section{Experiments}
\label{experiments:sec}

In this experimental section we try to validate our claims about time
complexity empirically, and explore some influences on scalability.
More specifically, we want to:
\begin{itemize}
\item validate the claim that when the localization assumption is exploited,
  induction time is linear in the number of examples (all other things
  being equal, i.e. we control for other influences on induction time such
  as the size of the tree)
\item study the influence of localization on induction time (by quantifying
  the amount of localization and investigating its effect on the time 
  complexity)
\item investigate how the induction time varies with the size of the data
  set in more practical situations (if we do {\em not} control other 
  influences; i.e. a larger data set may cause the learner to induce a more
  complex theory, which in itself has an effect on the induction time)
\end{itemize}

Before discussing the experiments themselves, we describe the data sets
that we have used.

\subsection{Description of the Data Sets}

\subsubsection{RoboCup Data Set}

This is a data set containing data about soccer games played by
software agents training for the RoboCup competition (Kitano
et al., 1997).
\nocite{Kitano97:proc}
It contains 88594 examples and is 100MB large.  Each example consists
of a description of the state of the soccer terrain as observed by one
specific player on a single moment.  This description includes the
identity of the player, the positions of all players and of the ball,
the time at which the example was recorded, the action the player
performed, and the time at which this action was executed.
Figure~\ref{RoboCupvb} shows one example.

\begin{figure}
\begin{verbatim}
begin(model(e71)).
   player(my,1,-48.804436,-0.16494742,339).
   player(my,2,-34.39789,1.0097091,362).
   player(my,3,-32.628735,-18.981379,304).
   player(my,4,-27.1478,1.3262547,362).
   player(my,5,-31.55078,18.985638,362).
   player(my,6,-41.653893,15.659259,357).
   player(my,7,-48.964966,25.731588,352).
   player(my,8,-18.363993,3.815975,362).
   player(my,9,-22.757153,32.208805,347).
   player(my,10,-12.914384,11.456045,362).
   player(my,11,-10.190831,14.468359,18).
   player(other,1,-4.242554,11.635328,314).
   player(other,2,0.0,0.0,0).
   player(other,3,-13.048958,23.604038,299).
   player(other,4,0.0,0.0,0).
   player(other,5,2.4806643,9.412553,341).
   player(other,6,-9.907758,2.6764495,362).
   player(other,7,0.0,0.0,0).
   player(other,8,0.0,0.0,0).
   player(other,9,-4.2189126,9.296844,339).
   player(other,10,0.4492856,11.43235,158).
   player(other,11,0.0,0.0,0).
   ball(-32.503292,0.81057936,362).
   mynumber(5).
   rctime(362).
   turn(137.4931640625).
   actiontime(362).
end(model(e71)).
\end{verbatim}
\caption{The Prolog representation of one example in the RoboCup data set.
A fact such as {\tt player(other,3,-13.048958,23.604038,299)} means that
player 3 of the other team was last seen at position (-13,23.6) at time
299.  A position of (0,0) means that that player has never been observed 
by the player that has generated this model.  The action performed currently
by this player is {\tt turn(137.4931640625)}: it is turning towards the ball.}
\label{RoboCupvb}
\end{figure}

While this data set would allow rather complicated theories to be
constructed, for our experiments the language bias was very simple and
consisted of a propositional language (only high-level commands are
learned).  This use of the data set reflects the
learning tasks considered up till now by the people who are using it,
see (Jacobs et al., 1998). \nocite{Jacobs98-ILP:proc}
This does not influence the validity of our results for
relational languages, because the propositions are defined by the
background knowledge and their truth values are computed at runtime,
so the query that is really executed is relational.  For instance,
the proposition {\tt have\_ball}, indicating whether some player of
the team has the ball in its possession, is computed from the
position of the player and of the ball.  

\subsubsection{Poker Data Sets}

The Poker data sets are artificially created data sets where each example
is a description of a hand of five cards, together with a name for the
hand (pair, three of a kind, \ldots).  The aim is to learn definitions for 
several poker concepts from a set of examples.  The classes that are 
considered here are {\tt nothing, pair, two\_pairs, three\_of\_a\_kind, 
full\_house} and {\tt four\_of\_a\_kind}.  This is, of course, a 
simplification of the real poker domain, where more classes exist and it is
necessary to distinguish between e.g. a pair of queens and a pair of kings; but
this simplified version suffices to illustrate the relevant topics
and keeps learning times sufficiently low to allow for reasonably extensive
experiments.

Figure~\ref{pokerdata} illustrates how one example in the poker domain
can be represented.  We have created the data sets for this domain
using a program that randomly generates examples for this domain.  The
advantage of this approach is its flexibility: it is easy to create
multiple training sets of increasing size, as well as an independent
test set.

\begin{figure}
\begin{verbatim}
begin(model(4)).
  card(7,spades).
  card(queen,hearts).
  card(9,clubs).
  card(9,spades).
  card(ace,diamonds).
  pair.
end(model(4)).
\end{verbatim}
\caption{An example from the Poker data set.}
\label{pokerdata}
\end{figure}

An interesting property of this data set is that some classes, e.g. 
{\tt four\_of\_a\_kind}, are very rare, hence a large data set is needed
to learn these classes (assuming the data are generated randomly).

\subsubsection{Mutagenesis Data Set}

The Mutagenesis dataset (Srinivasan et al., 1996)
\nocite{Srinivasan96:jrnl} is a classic benchmark in Inductive Logic
Programming.  The set that has been used most often in the literature
consists of 188 examples.  Each example describes a molecule.  Some of
these molecules are mutagenic (i.e., may cause DNA mutations), others
are not.  The task is to predict the mutagenicity of a molecule from
its description.

The data set is a typical ILP data set in that the example
descriptions are highly structured, and there is background knowledge
about the domain.  Several levels of background knowledge have been
studied in the literature (see again Srinivasan et al. (1996))\nocite{Srinivasan96:jrnl}; for
our experiments we have always used the simplest background knowledge,
i.e. only structural information about the molecules (the atoms and
bonds occurring in them) are available.

Figure~\ref{mutavb} shows a part of the description of one molecule.
\begin{figure}
\begin{verbatim}
begin(model(1)).
   pos.
   atom(d1_1,c,22,-0.117).
   atom(d1_2,c,22,-0.117).
   atom(d1_3,c,22,-0.117).
   atom(d1_4,c,195,-0.087).
   atom(d1_5,c,195,0.013).
   atom(d1_6,c,22,-0.117).
   (...)
   atom(d1_25,o,40,-0.388).
   atom(d1_26,o,40,-0.388).
   
   bond(d1_1,d1_2,7).
   bond(d1_2,d1_3,7).
   bond(d1_3,d1_4,7).
   bond(d1_4,d1_5,7).
   bond(d1_5,d1_6,7).
   bond(d1_6,d1_1,7).
   bond(d1_1,d1_7,1).
   bond(d1_2,d1_8,1).
   bond(d1_3,d1_9,1).
   (...)
   bond(d1_24,d1_19,1).
   bond(d1_24,d1_25,2).
   bond(d1_24,d1_26,2).
end(model(1)).
\end{verbatim}
\caption{The Prolog representation of one example in the Mutagenesis data set.
The {\tt atom} facts enumerate the atoms in the molecule.  For each atom its
element (e.g. carbon), type (e.g. carbon can occur in several configurations;
each type corresponds to one specific configuration) and partial charge.  The
{\tt bond} facts enumerate all the bonds between the atoms (the last
argument is the type of the bond: single, double, aromatic, etc.).  
{\tt pos} denotes that the molecule belongs to the positive class (i.e. is
mutagenic).}
\label{mutavb}
\end{figure}

\subsection{Materials and Settings}
\label{materials:sec}

All experiments were performed with the two implementations of \Tilde\ we
discussed: \Tilde{\em classic} and \Tilde{\em LDS}.  These programs are 
implemented in Prolog and run under the MasterProlog engine (formerly
named ProLog-by-BIM).  The hardware we used is a Sun Ultra-2 at 167 MHz,
running the Solaris system (except when stated otherwise).

Both \Tilde{\em classic} and \Tilde{\em LDS} offer the possibility to
precompile the data file.  We exploited this feature for all our experiments.
For \Tilde{\em LDS} this raises the problem that in order to load one
example at a time, a different object file has to be created for each
example (MasterProlog offers no predicates for loading only a part of
an object file).  This can be rather impractical.  For this reason
several examples are usually compiled into one object file; a
parameter called {\em granularity} ($G$) controls how many examples
can be included in one object file.

Object files are then loaded one by one by \Tilde{\em LDS}, which means
that $G$ examples at a time are loaded into main memory (instead of one).
Because of this, the granularity parameter can affect the
efficiency of \Tilde{\em LDS}.  This is investigated in our experiments.

By default, a value of 10 was used for $G$.

\subsection{Experiment 1: Time Complexity}

\subsubsection{Aim of the Experiment}

As mentioned before, induction of trees with \Tilde{\em LDS} should in
principle have a time complexity that is linear in the number of
examples.  With our first experiment we empirically test whether our
implementation indeed exhibits this property.  We also compare it
with other approaches where the locality assumption is exploited less
or not at all.

We distinguish the following approaches:

\begin{itemize}
\item loading all data at once in main memory without exploiting the locality
assumption (the standard ILP approach)
\item loading all data at once in main memory, exploiting the locality 
assumption; this is what \Tilde{\em classic} does
\item loading examples one at a time in main memory; this is what
\Tilde{\em LDS} does
\end{itemize}

To the best of our knowledge all ILP systems that do not learn from
interpretations follow the first approach (with the exception of a few
systems that access an external database directly instead of loading
the data into main memory, e.g. {\sc Rdt/db} (Morik and Brockhausen, 1997)
\nocite{Morik97:jrnl}; but these
systems still do not make a locality assumption).  We can easily
simulate this approach with \Tilde{\em classic} by specifying all
information about the examples as background knowledge.  For the background
knowledge no locality assumption can be made, since all
background knowledge is potentially relevant for each example.

The performance of a Prolog system that works with a large database
is improved significantly if indexes are built for the predicates.
On the other hand, adding indexes for predicates creates some overhead
with respect to the internal space that is needed, and a lot of
overhead for the compiler.  The MasterProlog system by
default indexes all predicates, but this indexing can be switched off.
We have performed experiments for the standard ILP approach both with
and without indexing (thus, the first approach in the above list is
actually subdivided into ``indexed'' and ``not indexed'').

\subsubsection{Methodology}

Since the aim of this experiment is to determine the influence of the number
of examples (and only that) on time complexity, we want to 
control as much as possible other factors that might also have an
influence.  We have seen in Section~\ref{scalability:sec} that
these other factors include the number of nodes $n$, the average number of
tests per node $t$ and the average complexity of performing one test on one
single example $c$.  $c$ depends on both the complexity of the queries
themselves and on the example sizes.

When varying the number of examples for our experiments, we want to keep
these factors constant.  This means that first of all the refinement
operator should be the same for all the experiments.  This
is automatically the case if the user does not change the refinement
operator specification (the {\tt rmode} facts) between consecutive 
experiments.  

The other factors can be kept constant by ensuring that the same tree
is built in each experiment, and that the average complexity of the
examples does not change.  In order to achieve this, we adopt the
following methodology.  We create, from a small data set, larger data
sets by including each single example several times.  By ensuring that
all the examples occur an equal number of times in the resulting data
set, the class distribution, average complexity of testing a query on
an example etc.\ are all kept constant.  In other words, all variation
due to the influence of individual examples is removed.  

Because the class distribution stays the same, the test that is
chosen in each node also stays the same.  This is necessary to ensure that
the same tree is grown, but not sufficient: the stopping criterion
needs to be adapted as well so that a node that cannot be split
further for the small data set is not split when using the larger data
set either.  In order to achieve this, the minimal number of examples
that have to be covered by each leaf (which is a parameter of \Tilde)
is increased proportionally to the size of the data set.

By following this methodology, the mentioned unwanted influences are 
filtered out of the results.

\subsubsection{Materials}

We used the Mutagenesis data set for this experiment.  Other materials
are as described in Section~\ref{materials:sec}.

\subsubsection{Setup of the Experiment}

Four different versions of \Tilde\ are compared:
\begin{itemize}
\item \Tilde{\em classic} without locality assumption, without indexing
\item \Tilde{\em classic} without locality assumption, with indexing
\item \Tilde{\em classic} with locality assumption
\item \Tilde{\em LDS}
\end{itemize}

The first three ``versions'' are actually the same version of \Tilde\ as
far as the implementation of the learning algorithm is concerned, but
differ in the way the data are represented and in the way the underlying
Prolog system handles them.

Each \Tilde\ version was first run on the original data set, then on
data sets that contain each original example $2^n$ times, with $n$
ranging from 1 to 9.
Table~\ref{mutasize} summarizes some properties of the data sets that were
obtained in this fashion.

For each run on each data set we have recorded the following:
\begin{itemize}
\item the time needed for the induction process itself (in CPU-seconds)
\item the time needed to compile the data (in CPU-seconds).  The different
systems compile the data in different ways (e.g. according to whether indexes
need to be built).  As compilation of the data need only be done
once, even if afterwards several runs of the induction system are done,
compilation time may seem less relevant.  Still, it is 
important to see how the compilation scales up, since it is
not really useful to have an induction method that scales linearly if it
needs a preprocessing step that scales super-linearly.
\end{itemize}

\subsubsection{Discussion of the Results}

\begin{figure}
\epsfxsize=9cm
\epsffile{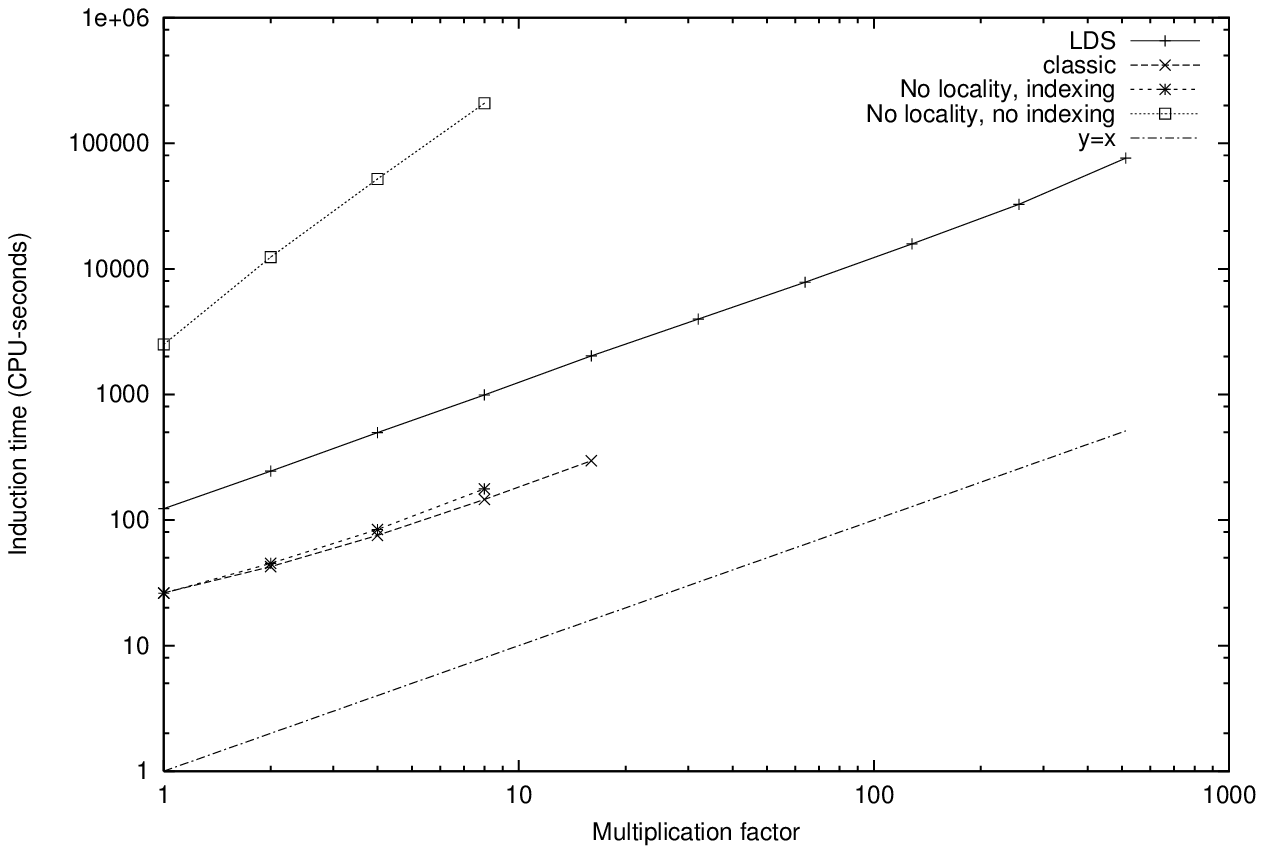}

\epsfxsize=9cm
\epsffile{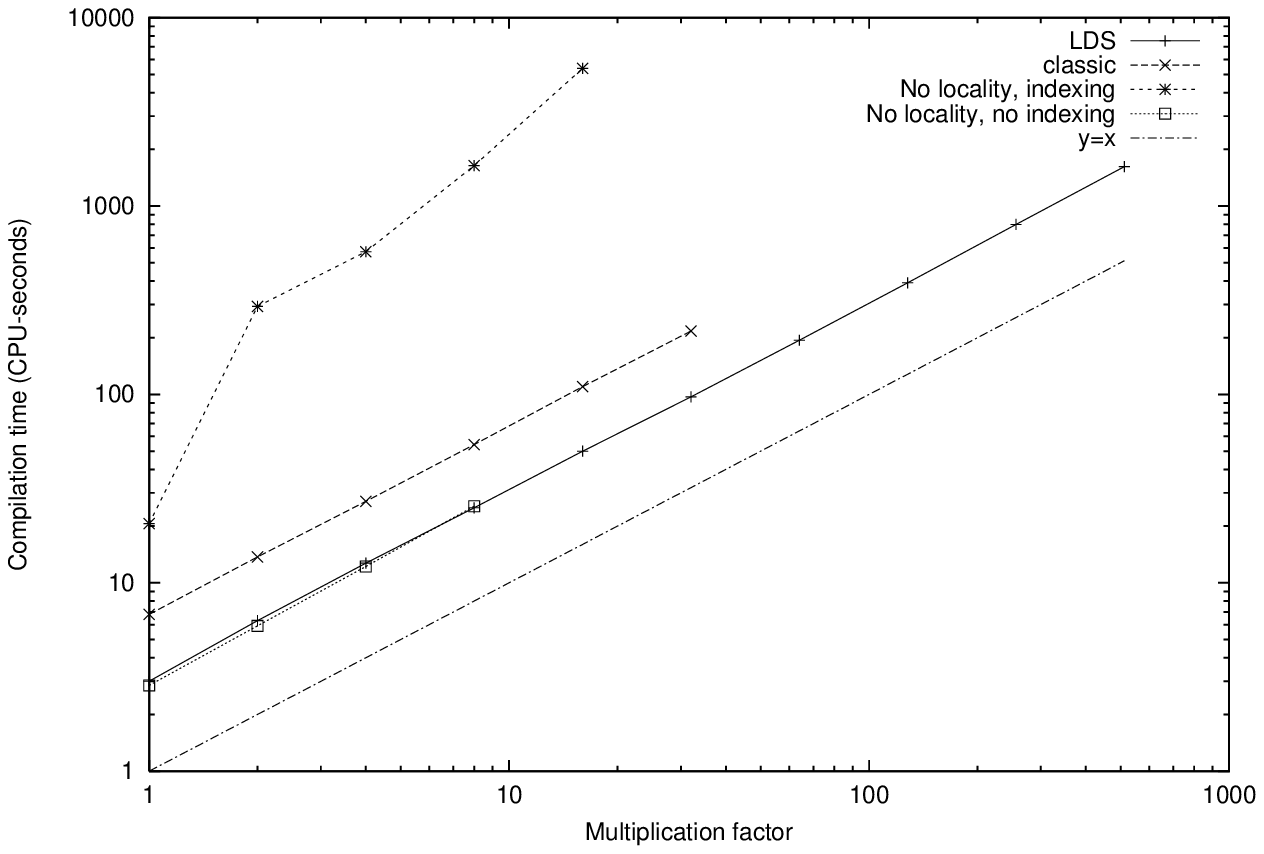}

\caption{Scaling properties of \Tilde{\em LDS} in terms of number of examples}
\label{mutafig}
\end{figure}

Tables~\ref{muta-LDS}, \ref{muta-classic}, \ref{muta-index} and
\ref{muta-noindex} give an overview of the time each \Tilde\ version
needed to induce a tree for each set, as well as the time it took to
compile the data into the correct format.  The results are shown
graphically in Figure~\ref{mutafig}.  Note that both the number of examples
and time are indicated on a logarithmic scale.  Care must be taken when
interpreting these graphs: a straight line does not indicate a linear
relationship between the variables.  Indeed, if $\log y = n * \log x$, then
$y = x^n$.  This means the slope of the line should be 1 in order to have
a linear relationship, while 2 indicates a quadratic relationship, and so on.
In order to make it easier to recognize a linear relationship (slope 1),
the function $y=x$ has been drawn on the graphs as a reference.

\begin{table}
\caption{Properties of the example sets}
\label{mutasize}
\begin{tabular}{lllll}
\hline
{\bf multiplication factor} & {\bf \#examples} & {\bf \#facts} & {\bf size (MB)}\\
\hline
1 & 188 & 10512 & 0.25 \\
2 & 376 & 21024 & 0.5 \\
4 & 752 & 42048 & 1 \\
8 & 1504 & 84096 & 2 \\
16 & 3008 & 168192 & 4 \\
32 & 6016 & 336384 & 8 \\
64 & 12032 & 672768 & 16 \\
128 & 24064 & 1,345,536 & 32 \\
256 & 48128 & 2,691,072 & 65 \\
512 & 96256 & 5,382,144 & 130  \\
\hline
\end{tabular}
\end{table}

\begin{table}
\caption{Scaling properties of \Tilde{\em LDS} in terms of the number of examples}
\label{muta-LDS}
\begin{tabular}{lll}
\hline
{\bf multiplication} & \multicolumn{2}{c}{{\bf time (CPU seconds)}}\\ 
{\bf factor} & {\bf induction} & {\bf compilation}\\
\hline
1 & 123 & 3 \\ 
2 & 245 & 6.3 \\ 
4 & 496 & 12.7 \\
8 & 992 & 25 \\ 
16 & 2026 & 50 \\ 
32 & 3980 & 97 \\ 
64 & 7816 & 194 \\ 
128 & 15794 & 391 \\ 
256 & 32634 & 799 \\ 
512 & 76138 & 1619 \\ 
\hline
\end{tabular}
\end{table}

\begin{table}
\caption{Scaling properties of \Tilde{\em classic} in terms of the number of examples}
\label{muta-classic}
\begin{tabular}{lll}
\hline
{\bf multiplication} & \multicolumn{2}{c}{{\bf time (CPU seconds)}} \\
{\bf factor} & {\bf induction} & {\bf compilation}\\
\hline
1 & 26.3 & 6.8 \\ 
2 & 42.5 & 13.7\\ 
4 & 75.4 & 27.1\\ 
8 & 148.7 & 54.2\\
16 & 296.1 & 110.1\\
32 & ?* & 217.1\\ 
\hline\\
\multicolumn{3}{l}{* Prolog engine failed to load the data}
\end{tabular}
\end{table}

\begin{table}
\caption{Scaling properties of \Tilde\ without locality assumption, with indexing, in terms of number of examples}
\label{muta-index}
\begin{tabular}{lll}
\hline
{\bf multiplication} & \multicolumn{2}{c}{{\bf time (CPU seconds)}} \\ 
{\bf factor} & {\bf induction} & {\bf compilation}\\
\hline
1 & 26.1 & 20.6 \\
2 & 45.2 & 293 \\ 
4 & 83.9 & 572 \\
8 & 176.7 & 1640 \\ 
16 & ?* & 5381 \\ 
32 & ?* & 18388 \\ 
\hline\\
\multicolumn{3}{l}{* Prolog engine failed to load the data}
\end{tabular}
\end{table}

\begin{table}
\caption{Scaling properties of \Tilde\ without locality assumption, without indexing, in terms of number of examples}
\label{muta-noindex}
\begin{tabular}{lll}
\hline
{\bf multiplication} & \multicolumn{2}{c}{{\bf time (CPU seconds)}} \\ 
{\bf factor} & {\bf induction} & {\bf compilation}\\
\hline
1 &   2501 & 2.85 \\ 
2 &  12385 & 5.91 \\ 
4 &  51953 & 12.21 \\
8 & 207966 & 25.47 \\
16 & ?* & 52.25 \\ 
32 &   & ?** \\ 
\hline\\
\multicolumn{3}{l}{* Prolog engine failed to load the data}\\
\multicolumn{3}{l}{** Prolog compiler failed to compile the data}
\end{tabular}
\end{table}

Note that only \Tilde{\em LDS} scales up well to large data sets.  The
other versions of \Tilde\ had problems loading or compiling the data
from a multiplication factor of 16 or 32 on.  

The graphs and tables show that induction time is linear in the number
of examples for \Tilde{\em LDS}, for \Tilde{\em classic} with locality,
and for \Tilde{\em classic} without locality but with
indexing.  For \Tilde{\em classic} without locality or indexing the
induction time increases quadratically with the number of examples.
This is not unexpected, as in this setting the time needed to run a
test on one single example increases with the size of the dataset.

With respect to compilation times, we note that all are linear in the size
of the data set, except \Tilde{\em classic} without locality and with indexing.
This is in correspondence with the fact that building an index for the
predicates in a deductive database is an expensive operation, super-linear
in the size of the database.

Furthermore, the experiments confirm that \Tilde{\em classic} with 
locality scales as well as \Tilde{\em LDS} with respect to time complexity,
but for large data sets runs into problems because it
cannot load all the data.

Observing that without indexing induction time increases
quadratically, and with indexing compilation time increases
quadratically, we conclude that the locality assumption is indeed
crucial to our linearity results, and that loading only a few examples
at a time in main memory makes it possible to handle much larger data
sets.

\subsection{Experiment 2: The Effect of Localization}

\subsubsection{Aim of the experiment}

In the previous experiment we studied the effect of the number of examples
on time complexity, and observed that this effect is different according
to whether the locality assumption is made.  In this experiment we do not
just distinguish between localized and not localized, but consider gradual
changes in localization, and thus try to quantify the effect of localization
on the induction time.

\subsubsection{Methodology}

We can test the influence of localization on the efficiency of
\Tilde{\em LDS} by varying the granularity parameter $G$ in \Tilde{\em
  LDS}.  $G$ is the number of examples that are loaded into main
memory at the same time.  Localization of information is stronger when
$G$ is smaller.  

The effect of $G$ was tested by running \Tilde{\em LDS} successively on
the same data set, under the same circumstances, but with different values
for $G$.  In these experiments $G$ ranged from 1 to 200.  For each value of 
$G$ both compilation and induction were performed ten times; the reported
times are the means of these ten runs.

\subsubsection{Materials}

We have used three data sets: a RoboCup data set with 10000
examples, a Poker data set containing 3000 examples, and the
Mutagenesis data set with a multiplication factor of 8 (i.e. 1504
examples).  The data sets were chosen to contain a sufficient
number of examples to make it possible to let $G$ vary over a relatively 
broad range, but not more (to limit the experimentation time).

Other materials are as described in Section~\ref{materials:sec}.

\subsubsection{Discussion of the Results}

\begin{figure}
\epsfxsize=9cm
\epsffile{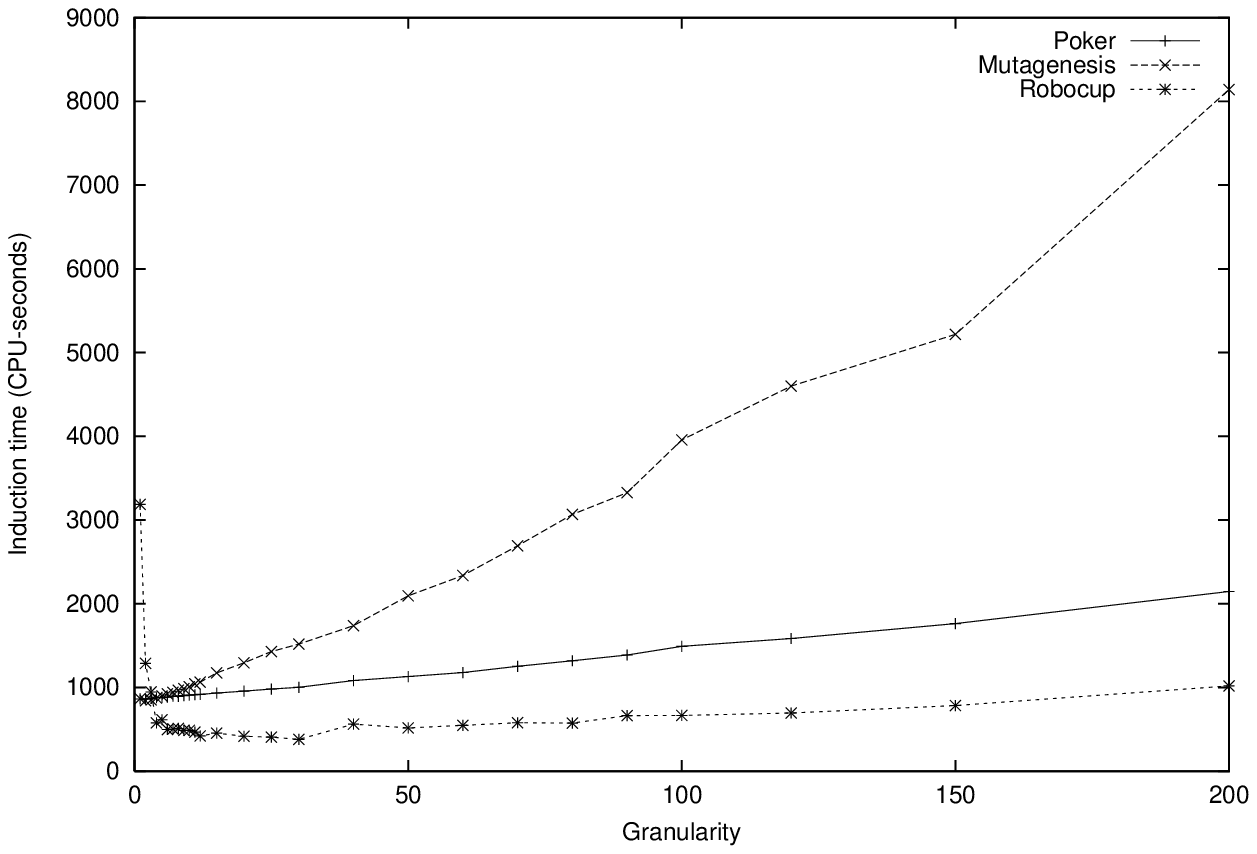}

\epsfxsize=9cm
\epsffile{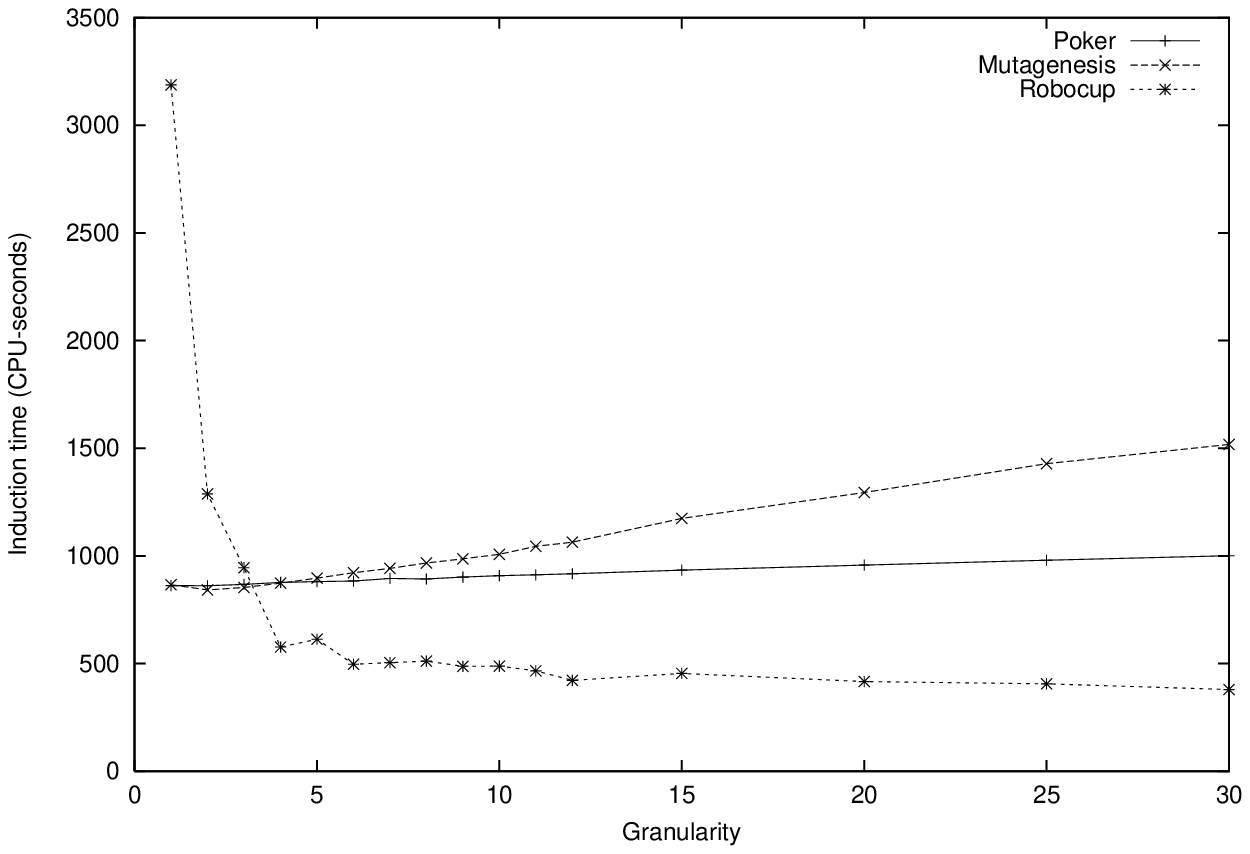}


\epsfxsize=9cm
\epsffile{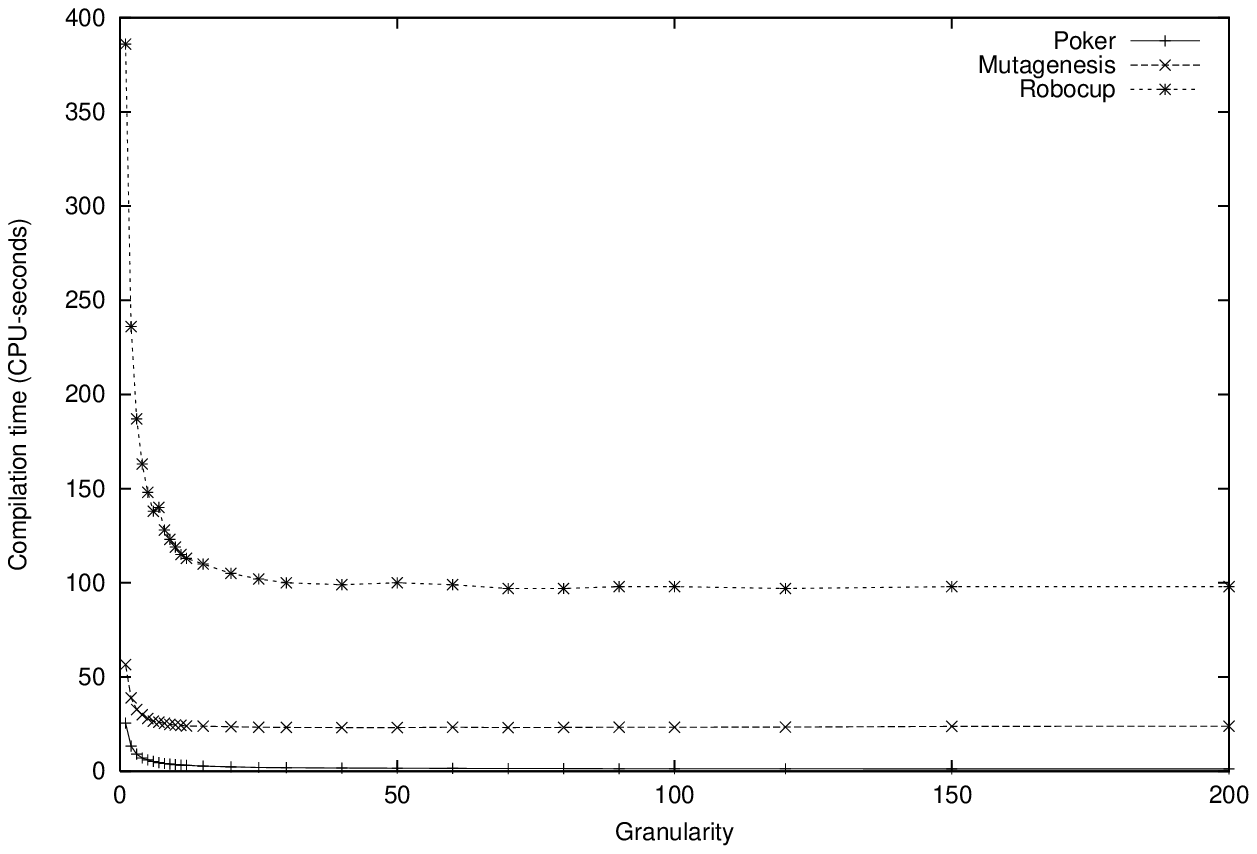}
\caption{The effect of granularity on induction time (full range, and zoomed
in on interval $[0-30]$) and compilation time}
\label{gran:fig}
\end{figure}

Induction times and compilation times are plotted versus granularity
in Figure~\ref{gran:fig}.  It can be seen from these plots that
induction time increases approximately linearly with granularity.  For 
very small
granularities, too, the induction time can increase.  We suspect that
this effect can be attributed to an overhead of disk access (loading
many small files, instead of fewer larger files).  A similar effect is
seen when we look at the compilation times: these decrease when the
granularity increases, but asymptotically approach a constant.  This
again suggests an overhead caused by compiling many small files
instead of one large file.  The fact that the observed effect is
smallest for Mutagenesis, where individual examples are larger,
increases the plausibility of this explanation.

This experiment clearly shows that the performance of \Tilde{\em LDS}
strongly depends on $G$, and that a reasonably small value for $G$ is
preferable.  It thus confirms the hypothesis that localization of
information is advantageous with respect to time complexity.

\subsection{Experiment 3: Practical Scaling Properties}

\subsubsection{Aim of the experiment}

With this experiment we want to measure how well \Tilde{\em LDS} scales up
in practice, without controlling any influences.  This means that the
tree that is induced is not guaranteed to be the same one or have the same
size, and that a natural variation is allowed with respect to the complexity
of the examples as well as the complexity of the queries.
This experiment is thus meant to mimic the situations that arise in practice.

Since different trees may be grown on different data sets, the quality
of these trees may differ.  We investigate this as well.

\subsubsection{Methodology}

The methodology we follow is to choose some domain and then create
data sets with different sizes for this domain.  \Tilde{\em LDS} is
then run on each data set, and for each run the induction time is
recorded, as well as the quality of the tree (according to different
criteria, see below).

\subsubsection{Materials}

Data sets from two domains were used: RoboCup and Poker.
These domains were chosen because large data sets were available for them.
For each domain several data sets of increasing size were created.

Whereas induction times have been measured on both data sets, predictive 
accuracy has been measured only for the Poker data set.  This was done
using a separate test set of 100,000 examples, which was the same for
all the hypotheses.

For the RoboCup data set interpretability of the hypotheses by domain
experts is the main evaluation criterion (because these theories are
used for verification of the behavior of agents, see (Jacobs {\em et
  al.}, 1998)\nocite{Jacobs98-ILP:proc}).

The RoboCup experiments have been run
on a SUN SPARCstation-20 at 100 MHz; for the Poker experiments a SUN Ultra-2
at 167 MHz was used.

\subsubsection{Discussion of the Results}

\begin{figure}
\epsfxsize=9cm
\epsffile{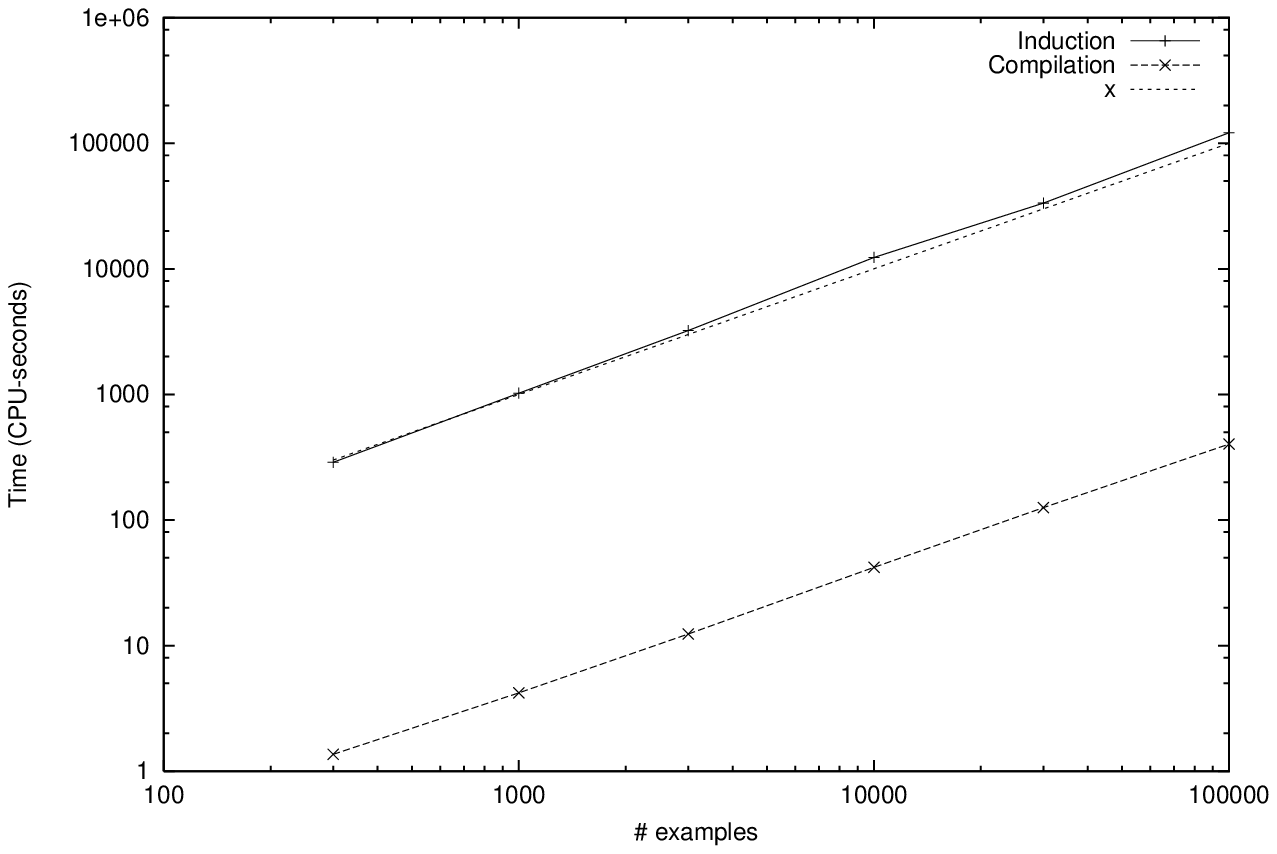}
\epsfxsize=9cm
\epsffile{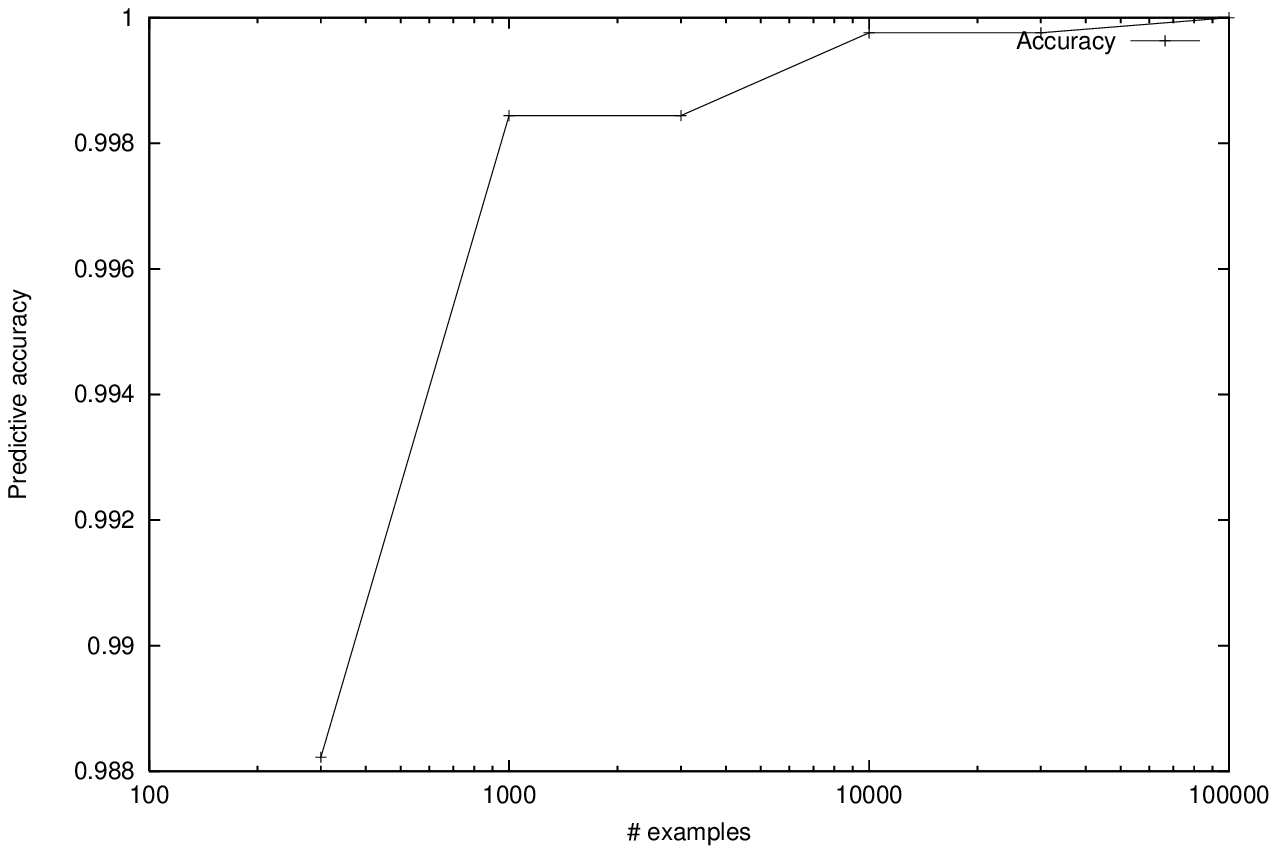}
\caption{Consumed CPU-time and accuracy of hypotheses produced
by \Tilde{\em LDS} in the Poker domain, plotted against the number of examples}
\label{pokertimes:fig}
\end{figure}

Table~\ref{pokertimes:tab} shows the consumed CPU-times in function
of the number of examples, as well as the predictive accuracy.  These
figures are plotted in Figure~\ref{pokertimes:fig}.  Note that the CPU-time
graph is again plotted on a double logarithmic scale.

\begin{table}
\caption{Consumed CPU-time and accuracy of hypotheses produced
by \Tilde{\em LDS} in the Poker domain}
\label{pokertimes:tab}
\begin{tabular}{llll}
\hline
{\bf \#examples} & {\bf compilation} & {\bf induction} & {\bf accuracy}\\
& {\bf (CPU-seconds)} & {\bf (CPU-seconds)} & \\
\hline
300    & 1.36 & 288 & 0.98822\\
1000   & 4.20 & 1021 & 0.99844\\
3000   & 12.36 & 3231 & 0.99844\\
10000  & 41.94 & 12325 & 0.99976\\
30000  & 125.47 & 33394 & 0.99976\\
100000 & 402.63 & 121266 & 1.0\\
\hline
\end{tabular}
\end{table}

\begin{table}
\caption{Consumed CPU-time of hypotheses produced by \Tilde{\em LDS} 
in the RoboCup domain}
\label{robotimes:tab}
\begin{tabular}{llll}
\hline
{\bf \#examples} & {\bf compilation} & {\bf induction} \\
\hline
10000 & 274 & 1448 $\pm$ 44\\
20000 & 522 & 4429 $\pm$ 83\\
30000 & 862 & 7678 $\pm$ 154\\
40000 & 1120 & 9285 $\pm$ 552\\
50000 & 1302 & 6607 $\pm$ 704\\
60000 & 1793 & 13665 $\pm$ 441\\
70000 & 1964 & 29113 $\pm$ 304\\
80000 & 2373 & 28504 $\pm$ 657\\
88594 & 2615 & 50353 $\pm$ 3063\\
\hline
\end{tabular}
\end{table}

\begin{figure}
\epsfxsize=9cm
\epsffile{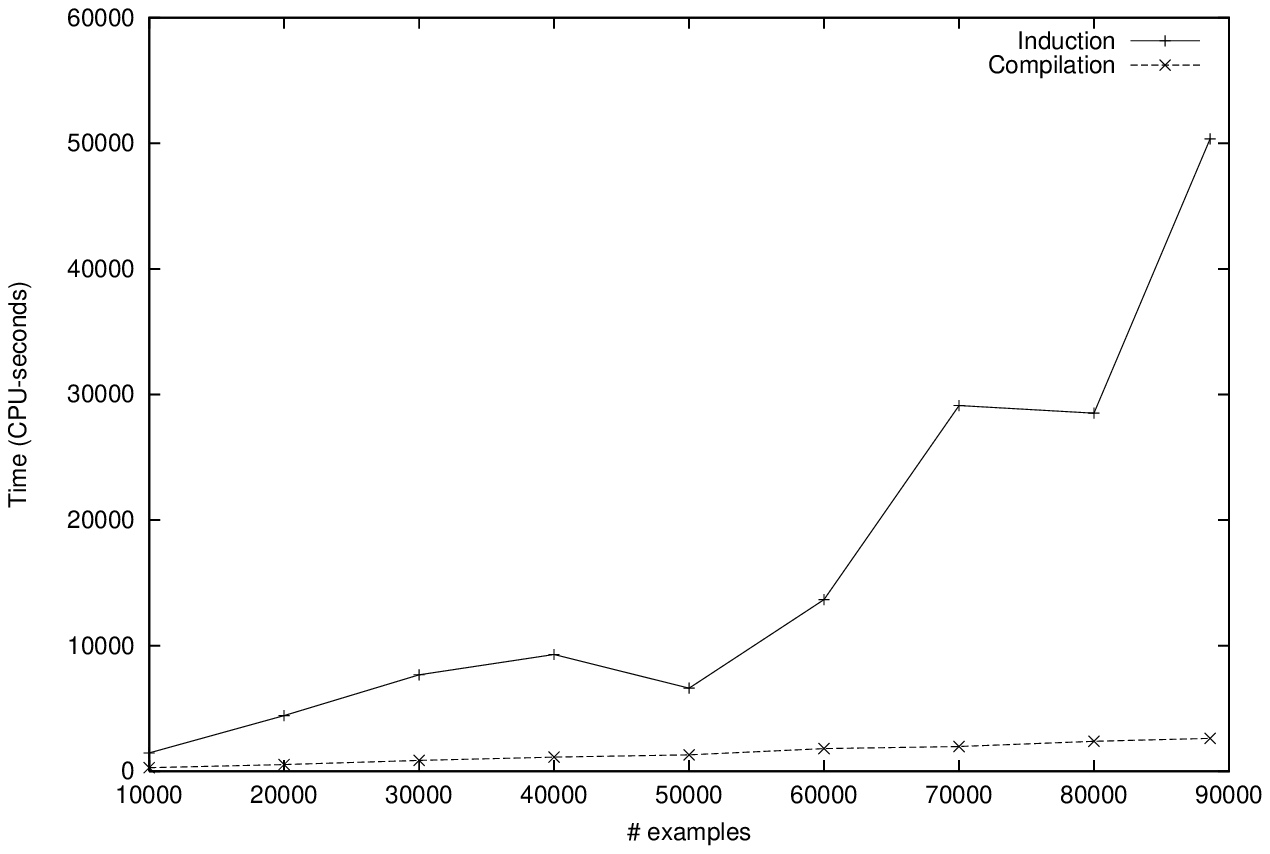}
\caption{Consumed CPU-time for \Tilde{\em LDS} in the RoboCup domain, plotted against the number of examples}
\label{robotimes:fig}
\end{figure}

With respect to accuracy, the Poker hypotheses show the expected behavior:
when more data are available, the hypotheses can predict very rare classes
(for which no examples occur in smaller data sets), which results in higher
accuracy.

The graphs further show that in the Poker domain, \Tilde{\em LDS} scales up
linearly, even though more accurate (and slightly more complex) theories
are found for larger data sets.  

In the RoboCup domain, the induced hypotheses were the same for all
runs except the 10000 examples run.  In this single case the
hypothesis was more simple and, according to the domain expert, less
informative than for the other runs.  This suggests that in this domain
a relatively small set of examples (20000) suffices to learn from.

It is harder to see how \Tilde{\em LDS} scales up for the RoboCup
data.  Since the same tree is returned in all runs except the 10000
examples run, one would expect the induction times to grow linearly.
However, the observed curve does not seem linear, although it does not
show a clear tendency to be super-linear either.  Because large
variations in induction time were observed, we performed these runs 10
times; the estimated mean induction times are reported together with
their standard errors.  The standard errors alone cannot explain the
observed deviations, nor can variations in example complexity (all
examples are of equal complexity in this domain).

A possible explanation is the fact that the Prolog engine performs a number of
tasks that are not controlled by \Tilde, such as garbage collection.
In specific cases, the Prolog engine may perform many garbage
collections before expanding its memory space (this happens when the amount of
free memory after garbage collection is always just above some
threshold), and the time needed for these garbage collections is
included in the measured CPU-times.  The MasterProlog engine is known
to sometimes exhibit such behavior.

In order to sort this out, \Tilde{\em LDS} would have to be reimplemented
in a lower-level language than Prolog, where one has full control over all
computations that occur.  Such a reimplementation is planned.

Due to the domain-dependent character of these complexity results, one
should be careful when generalizing them; it seems
safe to conclude, however, that the linear scaling property has
at least a reasonable chance of occurring in practice.

\section{Related Work}

Our work is closely related to efforts in the propositional learning
field to increase the capability of machine learning systems to handle
large databases.  It has been influenced more specifically by a
tutorial on data mining by Usama Fayyad, in which the work of Mehta
and others was mentioned (Mehta et al., 1996; Shafer {\em et
  al.}, 1996).\nocite{Mehta96:proc,Shafer96:proc} They were the first
to propose the level-wise tree building algorithm we adopted, and to
implement it in the SLIQ (Mehta et al., 1996) and SPRINT (Shafer
et al., 1996)\nocite{Shafer96:proc} systems.  The main
difference with our approach is that SLIQ and SPRINT learn from one
single relation, while \Tilde{\em LDS} can learn from multiple
relations.

Related work inside ILP includes the {\sc Rdt/db} system (Morik and
Brockhausen, 1997)\nocite{Morik97:jrnl}, which presents the first
approach to coupling an ILP system with a relational database
management system (RDBMS).  Being an ILP system, {\sc Rdt/db} also learns
from multiple relations.  The approach followed is that a logical test
that is to be performed is converted into an SQL query and sent to an
external relational database management system.  This approach is
essentially different from ours, in that it exploits as much as
possible the power of the RDBMS to efficiently evaluate queries.
Also, there is no need for preprocessing the data.  Disadvantages are
that for each query an external database is accessed, which is
relatively slow, and that it is less flexible with respect to
background knowledge.  Furthermore, to obtain good performance complex
modifications to the RDBMS system (tailoring it towards data mining)
are needed.  Preliminary experiments with coupling {\sc Claudien} and
\Tilde\ to an Oracle RDBMS confirmed these claims and caused us to
abandon such an approach.

We also mention the KEPLER system (Wrobel et al., 1996) 
\nocite{Wrobel96:proc}, a data mining
tool that provides a framework for applying a broad range of data
mining systems to data sets; this includes ILP systems.  KEPLER was
deliberately designed to be very open, and systems using the learning
from interpretations setting can be plugged into it as easily as other
systems.

At this moment few systems use the learning from interpretations
setting (De Raedt and Van Laer, 1995; De Raedt and Dehaspe, 1997; Dehaspe
and De Raedt, 1997).
\nocite{Deraedt95-ALT:proc,DeRaedt97-ML:jrnl,Dehaspe97a:proc}
Of these the research described in (Dehaspe and De Raedt, 1997) 
\nocite{Dehaspe97a:proc} (the {\sc Warmr}
system: finding association rules over multiple relations; see also
Dehaspe and Toivonen's contribution in this issue) is most closely
related to the work described in this paper, in the sense that there,
too, an effort was made to adapt the system for large databases.  The
focus of that text is not on the advantages of learning from
interpretations in general, however, but on the power of first
order association rules.

More loosely related work inside ILP would include all efforts to make
ILP systems more efficient.  Since most of this work concerns ILP
systems that work in the classical ILP setting, the ways in which this
is done usually differ substantially from what we describe in this
paper.  For instance, the well-known ILP system Progol (Muggleton,
1995) \nocite{Muggleton95-NGC:jrnl} has recently been extended with
caching and other efficiency improvements (Cussens, 1997).
\nocite{Cussens97:proc} Other directions are the use of sampling
techniques and stochastic methods, such as proposed by, e.g., Srinivasan
(1999) and Sebag (1998).

Finally, the \Tilde\ system is related to other systems that induce
first order decision trees, such as the {\sc Struct} system (Watanabe and 
Rendell, 1991)
\nocite{Watanabe91:proc} 
(which uses a less explicitly logic-based approach) and 
the regression tree learner SRT (Kramer, 1996).\nocite{Kramer96:proc}

\section{Conclusions}

We have argued and demonstrated empirically that the use of ILP is not
limited to small databases, as is often assumed.  Mining databases 
of a hundred megabytes was shown to be feasible, and this does not seem to
be a limit.

The positive results that have been obtained are due mainly to the use of
the {\em learning from interpretations} setting, which is
more scalable than the classical ILP setting and makes the
link with propositional learning more clear.  This means that a lot of
results obtained for propositional learning can be extrapolated to
learning from interpretations.  We have discussed a number of such upgrades,
using the \Tilde{\em LDS} system as an illustration.
The possibility to upgrade the work by Mehta et al. (1996) has turned
out to be crucial for handling large data sets.  It is not clear how the
same technique could be incorporated in a system using the classical ILP 
setting.

Although we obtained specific results only for a specific kind of data
mining (induction of decision trees), the results are generalizable
not only to other approaches within the classification context (e.g. rule
based approaches) but also to other inductive tasks within the
learning from interpretations setting, such as clustering, regression and
induction of association rules.

\subsubsection*{Acknowledgements}

Nico Jacobs and Hendrik Blockeel are supported by the Flemish Institute
for the Promotion of Scientific and Technological Research in the Industry
(IWT).  Luc De Raedt is supported by the Fund for Scientific Research,
Flanders.  This work is also part of the European Community Esprit
project no. 20237, Inductive Logic Programming 2.

The authors thank Luc Dehaspe, Kurt Driessens, H{\'e}l{\`e}ne Legras and
Jan Ramon for proofreading this text, as well as the anonymous reviewers
and Sa{\v s}o D{\v z}eroski for their very valuable comments on an 
earlier draft.

\bibliographystyle{plain}
\bibliography{.mlbib}

\end{document}